%% file: main.tex
\documentclass[10pt,twocolumn,letterpaper]{article}

\usepackage{cvpr}
\usepackage{times}
\usepackage{epsfig}
\usepackage{graphicx}
\usepackage{caption}
\usepackage{amsmath}
\usepackage{amssymb}
\usepackage{tabularx}
\usepackage[normalem]{ulem}
\usepackage{multirow}
\usepackage{xcolor}
\usepackage{float}
\usepackage{placeins}
\usepackage{abstract} 
\usepackage{xr}

\definecolor{road}{RGB}{128,64,128}
\definecolor{sidewalk}{RGB}{244,35,232}
\definecolor{building}{RGB}{70,70,70}
\definecolor{wall}{RGB}{102,102,156}
\definecolor{fence}{RGB}{190,153,153}
\definecolor{pole}{RGB}{153,153,153}
\definecolor{light}{RGB}{250,170,30}
\definecolor{sign}{RGB}{220,220,0}
\definecolor{vegetation}{RGB}{107,142,35}
\definecolor{terrain}{RGB}{152,251,152}
\definecolor{sky}{RGB}{70,130,180}
\definecolor{person}{RGB}{220,20,60}
\definecolor{rider}{RGB}{255,0,0}
\definecolor{car}{RGB}{0,0,142}
\definecolor{truck}{RGB}{0,0,70}
\definecolor{bus}{RGB}{0,60,100}
\definecolor{train}{RGB}{0,80,100}
\definecolor{motorcycle}{RGB}{0,0,230}
\definecolor{bicycle}{RGB}{119,11,32}

\usepackage[normalem]{ulem}

\usepackage[pagebackref=true,breaklinks=true,letterpaper=true,colorlinks,bookmarks=false]{hyperref}

\cvprfinalcopy 

\title{Synscapes: A Photorealistic Synthetic Dataset for Street Scene Parsing}

\author{Magnus Wrenninge$^{1,}$\thanks{magnus@7dlabs.com}\quad Jonas Unger$^{1,2,}$\thanks{jonas@7dlabs.com}\\\vspace{-3mm}\quad\\$^{1}$7D Labs \qquad $^{2}$Link\"oping University, Sweden}

\begin{document}
\twocolumn[{%
\renewcommand\twocolumn[1][]{#1}%
\maketitle
\begin{center}
    \includegraphics[width=0.99\textwidth,trim={0 1cm 0 1cm}, clip]{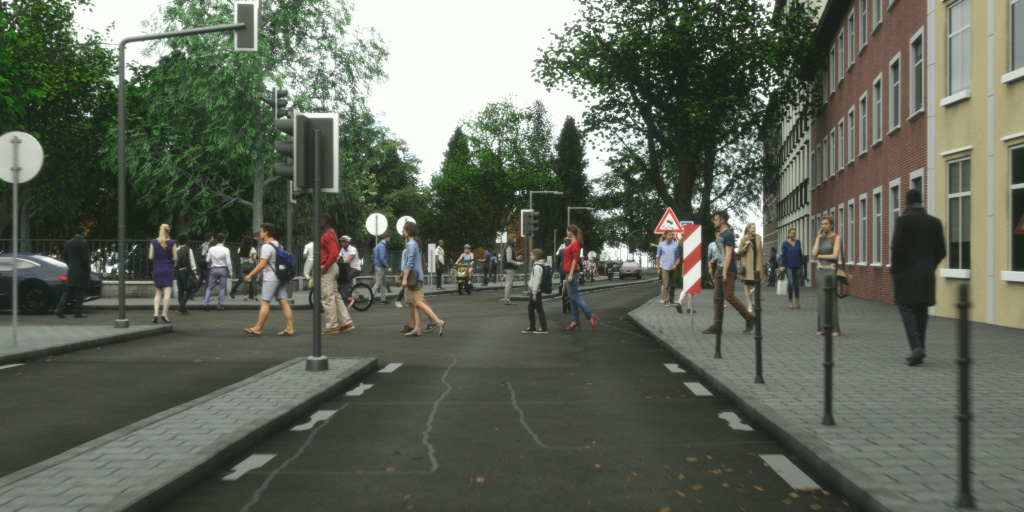}
    \captionof{figure}{Example image from Synscapes. 
    \label{fig:teaser}}
\end{center}%
}]

\saythanks 



\begin{abstract}
\noindent We introduce Synscapes -- a synthetic dataset for street scene parsing created using photorealistic rendering techniques, and show state-of-the-art results for training and validation as well as new types of analysis. 

We study the behavior of networks trained on real data when performing inference on synthetic data: a key factor in determining the equivalence of simulation environments. We also compare the behavior of networks trained on synthetic data and evaluated on real-world data.

Additionally, by analyzing pre-trained, existing segmentation and detection models, we illustrate how uncorrelated images along with a detailed set of annotations open up new avenues for analysis of computer vision systems, providing fine-grain information about how a model's performance changes according to factors such as distance, occlusion and relative object orientation.
\end{abstract}


\section{Introduction}

\begin{figure*}
\includegraphics[width=0.249\textwidth]{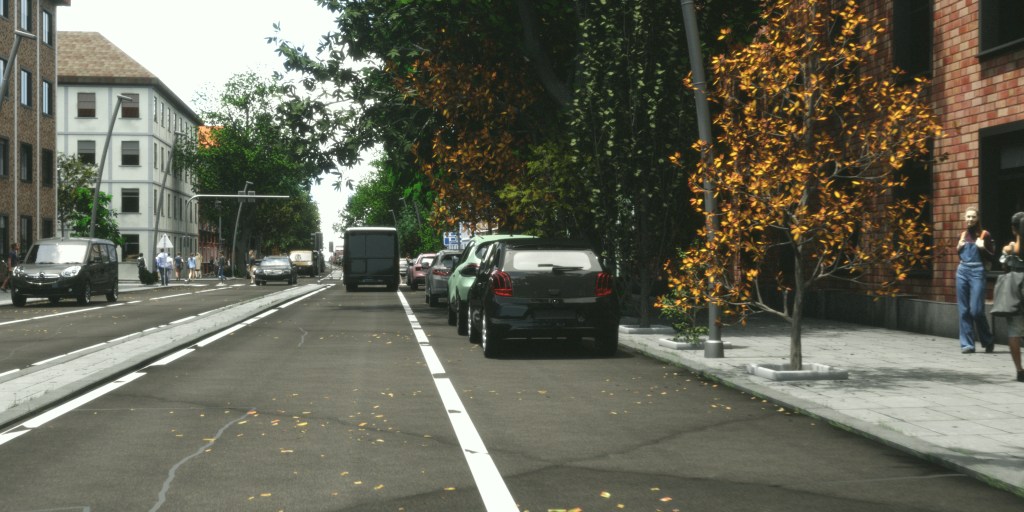}\includegraphics[width=0.249\textwidth]{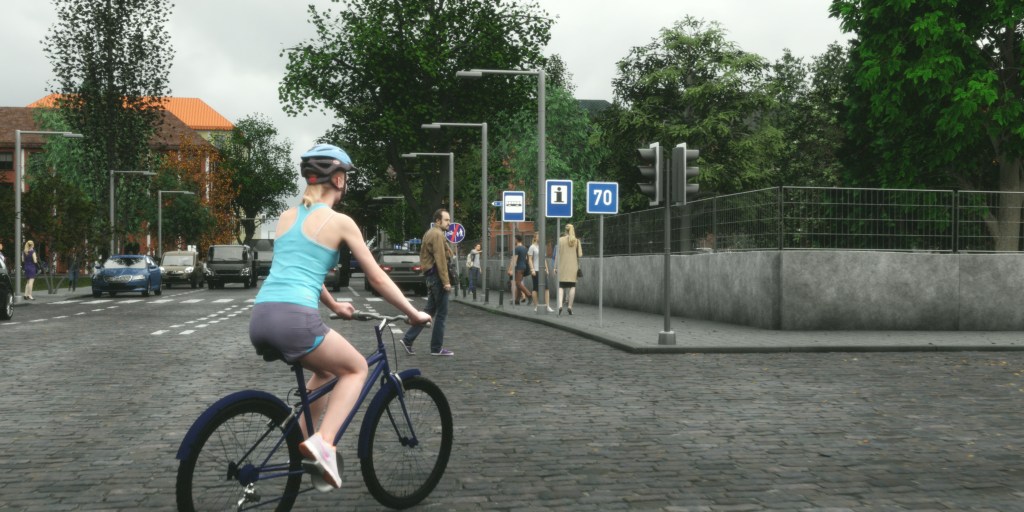}\includegraphics[width=0.249\textwidth]{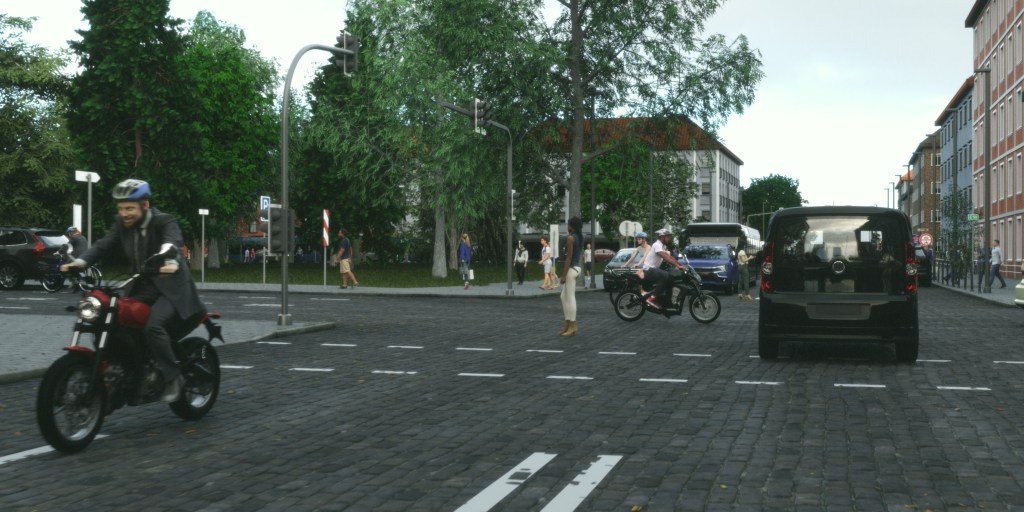}\includegraphics[width=0.249\textwidth]{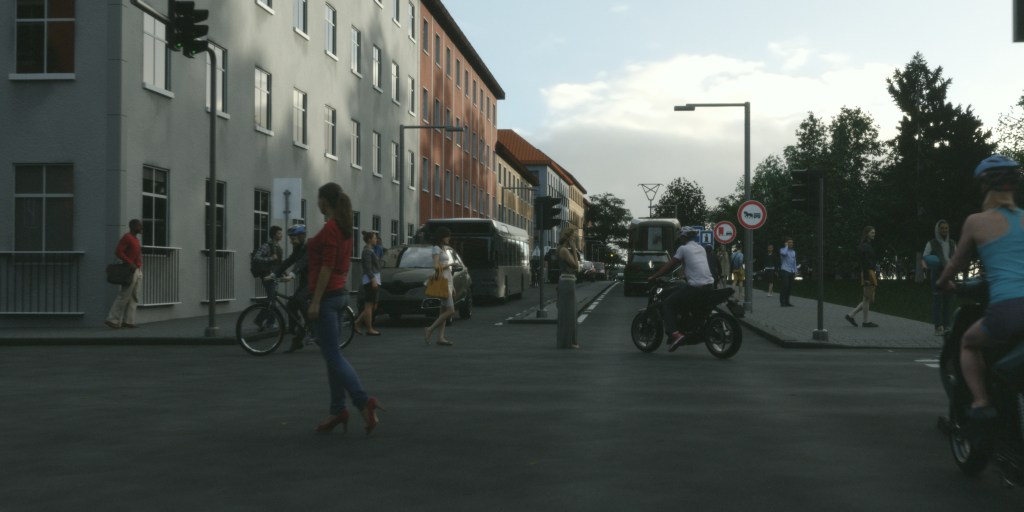}
\includegraphics[width=0.249\textwidth]{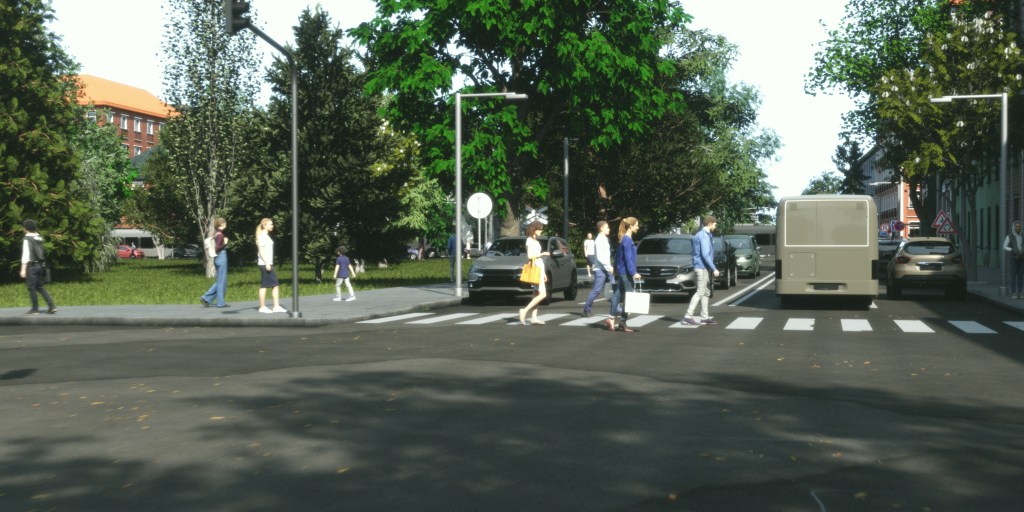}\includegraphics[width=0.249\textwidth]{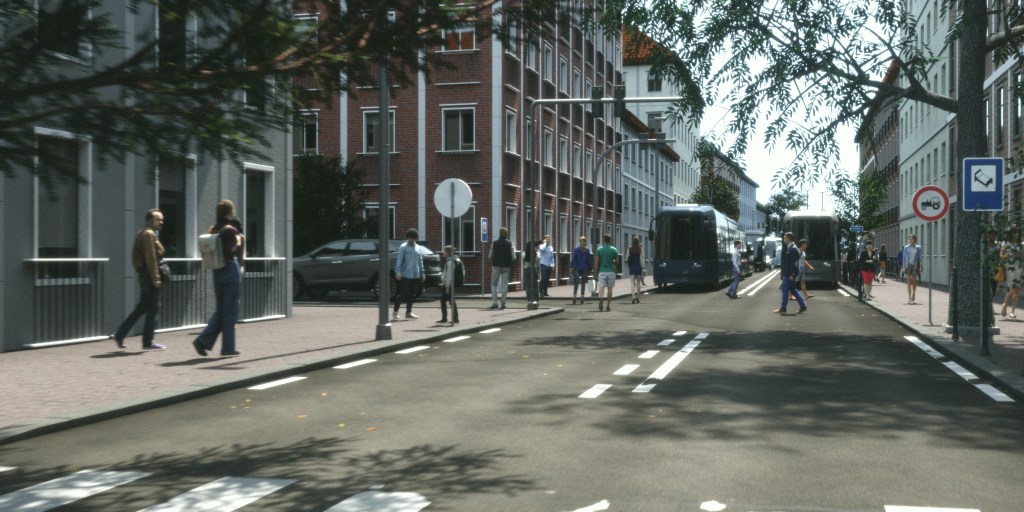}\includegraphics[width=0.249\textwidth]{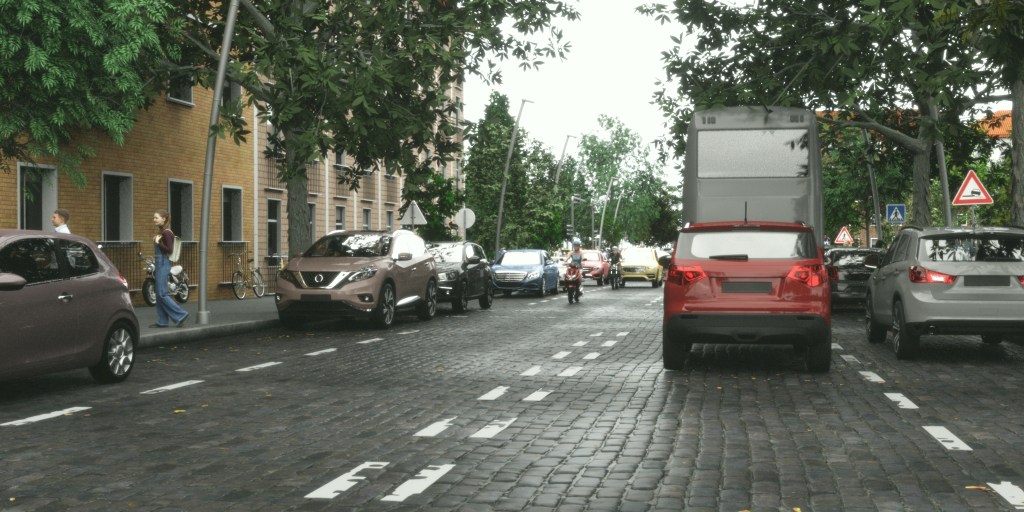}\includegraphics[width=0.249\textwidth]{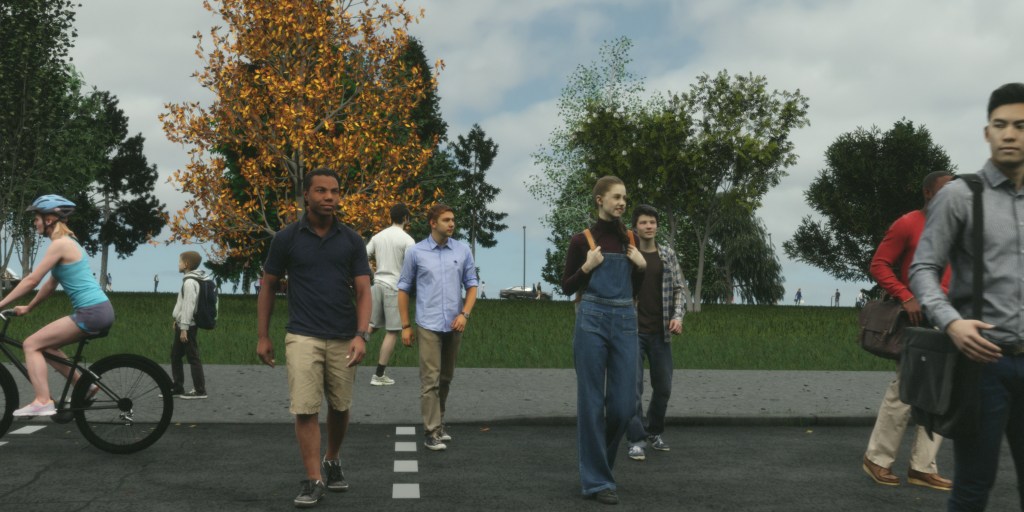}
\caption{The scenario for each image in Synscapes is unique, and the dataset encompasses a wide variety of street scene configurations and environmental factors.\label{fig:examples}}
\end{figure*}

\noindent There is a range of data that can be considered synthetic in the machine learning context. For example, the common technique of using \textit{augmentation} to create variations in data during training is a light-weight form of synthesis. On the other end of the scale we have approaches for creating data entirely by artificial means. 
Synscapes, along with several preceding synthetic datasets for computer vision tasks (and for street scene parsing in particular) are examples of the latter.

Synthetic datasets for computer vision tasks generally use \textit{computer graphics} in order to create images that can be used for training and validation of machine learning systems. Most \cite{Richter2016,Ros_2016_CVPR} tend to use \textit{game engines} to render the final images, but some \cite{blasinski2018optimizing} have also used offline, physically based rendering, commonly employed in visual effects production and animation, to produce their datasets.

Synthetic data generation methods generally highlight the low cost of producing the data as the main benefit, but this focus obscures some of the perhaps more important with synthetic data. Namely, the ability to produce arbitrary amounts of data from arbitrary probability distributions with arbitrarily detailed annotations. And in this sense, not all approaches to generating synthetic data are equal.

\section{Previous synthetic street scene datasets}

\noindent Virtual KITTI \cite{gaidon2016virtual} closely recreates parts of the KITTI \cite{geiger2013vision} dataset at a high level: buildings and individual actors are placed identically and the field of view matches. But the complexity and realism are both low.

Synthia \cite{Ros_2016_CVPR} provides images from a virtual world created within the Unity framework. It was one of the earliest work showing that synthetic data could be assembled by leveraging off-the-shelf assets which were recombined to create relevant street scenarios.

Richter et al.~\cite{Richter2016} used Grand Theft Auto V to create a much more complex environment than Synthia, but still relied on simplistic geometry and non-photorealistic real-time rendering. Although an apparent easy-to-access source of data, the legality of using existing games is questionable, and it obscures the amount of technical and artistic work required to create such virtual worlds, which often ranges in the hundreds of person-years.

Recently, Richter et al.~\cite{Richter_2017} released the Playing for Benchmarks dataset, which extended the use of GTA V with a larger set of images and wider range of annotations.

\section{Synscapes overview}
\label{sec:overview}
\begin{figure}
{\centering
\includegraphics[width=0.7\linewidth]{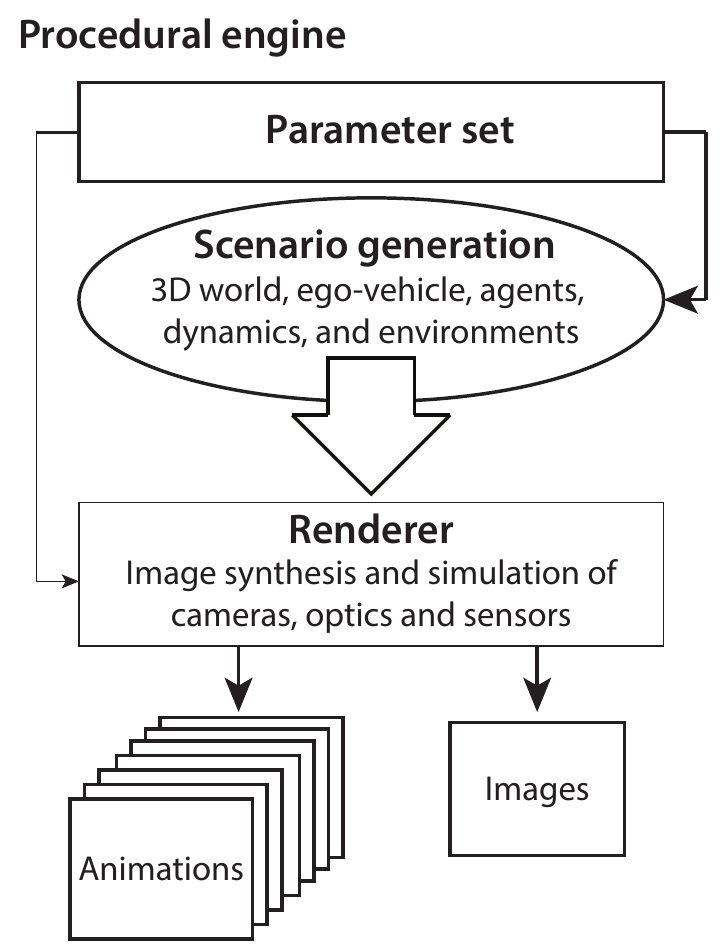}
\caption{Overview of the procedural approach for scenario and image generation. Each image is defined by a scenario created from a specific sampling of the generating parameters.
The benefit of the procedural approach is that it enables full control over the parameter sampling, scenario generation and rendering without manual work. \label{fig:overview}}
}
\end{figure}

\noindent Synscapes is created with an end-to-end approach to realism, accurately capturing the effects of everything from illumination by sun and sky, to the scene's geometric and material composition, to the optics, sensor and processing of the camera system. The images in the dataset do not follow a driven path through a single virtual world. Instead, an entirely unique scene is procedurally generated for each of the twenty-five thousand images. As a result, the dataset contains a wide range of variations and unique combinations of features. 

The procedural engine, illustrated in Figure~\ref{fig:overview} and described in more detail in a previous paper~\cite{tsirikoglolu2017procedural}, parameterizes all aspects of the 3D world generation and the image synthesis, and enables fully automated production of datasets. A \emph{scenario parameter} controls e.g. the amount of cars or pedestrians, the width of the road, the road surface material, the time of day, or weather conditions. A \emph{scenario} is an instantiation of a 3D world (the ego-vehicle, agents, etc.) defined by the sampling of a point in the high dimensional space of the scenario parameters. Each parameter in our system is coupled with a distribution such that each sample can be drawn in a statistically meaningful way. The scenario defines the input to the rendering engine responsible for simulation of sensors and optics. A scenario can be used for rendering both animations and single images. Since Synscapes is meant for experiments in training and validation, it consists of only single images, generated from parameter samples that ensure a wide coverage and variation of classes and features.

The images were rendered using unbiased path tracing \cite{Kajiya:1986:RE:15886.15902}: the same physically based rendering technique that powers high-end visual effects in the film industry. Light transport is calculated using radiometric properties from the sun and the sky, modeling the light's interaction with surfaces using physically based reflectance models, ensuring that each image is representative of the real world. Additionally, the effects of light scattering in the camera optics is modeled using a long-tail point spread function (PSF), and effects related to the imaging sensor such as readout noise, camera response function (CRF) and color characteristics are also simulated. 

As the name implies, Synscapes was designed to be similar in structure and content to the Cityscapes dataset \cite{Cordts2016Cityscapes}, and it includes all 19 of its training classes for semantic segmentation.

\subsection{RGB camera images}

\noindent Synscapes consists of 25,000 RGB images in PNG format at $1440 \times 720$ resolution, stored in the \texttt{img/rgb} subdirectory. In order to facilitate easier training on architectures configured for the Cityscapes dataset, we also provide the same images at $2048 \times 1024$ resolution in the \texttt{img/rgb-2k} subdirectory. For the latter, the sensor simulation was executed at the higher resolution, so that individual pixels carry the appropriate noise profile, rather than up-sampled noise. The images are sequentially numbered with no padding, ranging from \texttt{1.png} to \texttt{25000.png}.

\subsection{Annotation images} 

\noindent Each image is annotated with class, instance and depth information. The class annotation is stored as a single-channel PNG and uses the Cityscapes class id convention\footnote{\href{https://github.com/mcordts/cityscapesScripts/blob/master/cityscapesscripts/helpers/labels.py}{https://github.com/mcordts/cityscapesScripts}}. The instance images encode the instance id in the RGB channels such that the original id can be recovered according to 
\begin{equation*}
R + 256 \cdot G + 256^2 \cdot B.
\end{equation*}

It should be noted that actors that are more than 99\% occluded may be removed from the metadata file, but can still have small numbers of visible pixels in the RGB images.

The per-pixel depth values are stored in the floating point OpenEXR format, recording the planar depth (i.e. the z-depth component) of each pixel in meters, not the physical distance.

\subsection{Metadata}

\noindent Metadata associated with each image is stored in the \texttt{meta} subdirectory, with a single JSON file corresponding to each RGB image.  Three types of metadata are provided: scene metadata, which describes properties of the scene as a whole; camera metadata, describing the intrinsics and extrinsics of the sensor; and instance metadata, which provides details on the individual actors in each image. These are described in more detail in the paper's appendix. As a result of a particular quirk in how Cityscapes and other datasets classify bicyclists and motorcyclists, the instance considers the union of both the rider and its mount, with the class determined by the vehicle type. The class segmentation image, however, distinguishes between the rider and the vehicle, as expected.

\subsection{Distribution of metadata parameters}

\begin{figure}
\includegraphics[width=0.499\columnwidth]{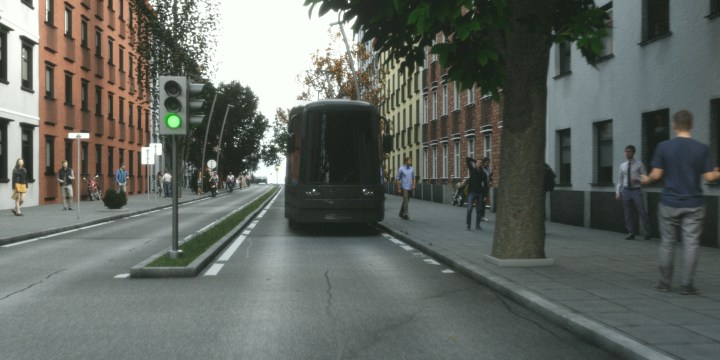}\includegraphics[width=0.499\columnwidth]{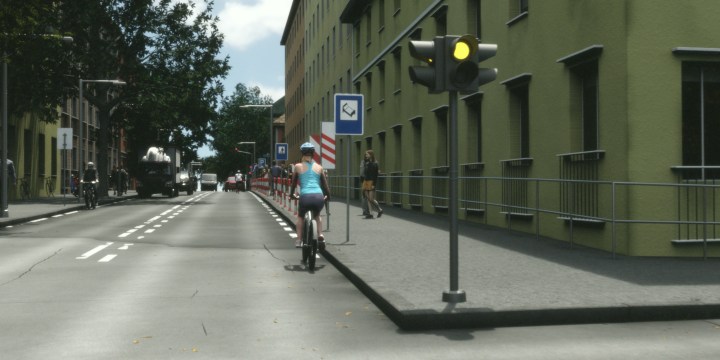}
\includegraphics[width=0.499\columnwidth]{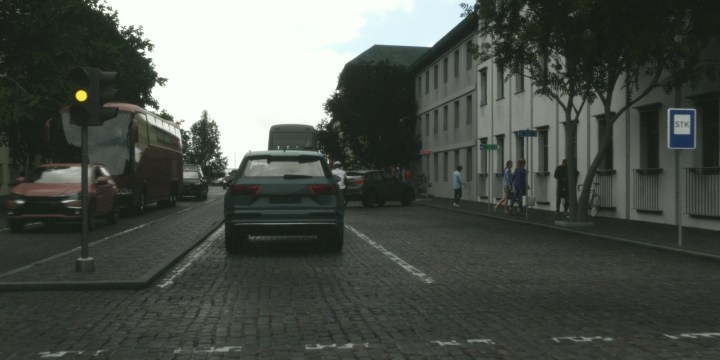}\includegraphics[width=0.499\columnwidth]{./images/2921-half.jpg}
\includegraphics[width=0.499\columnwidth]{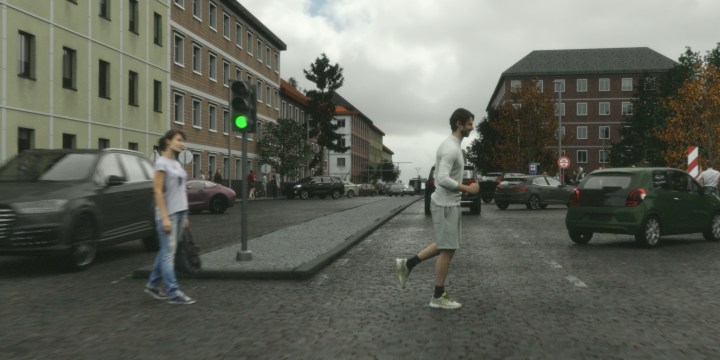}\includegraphics[width=0.499\columnwidth]{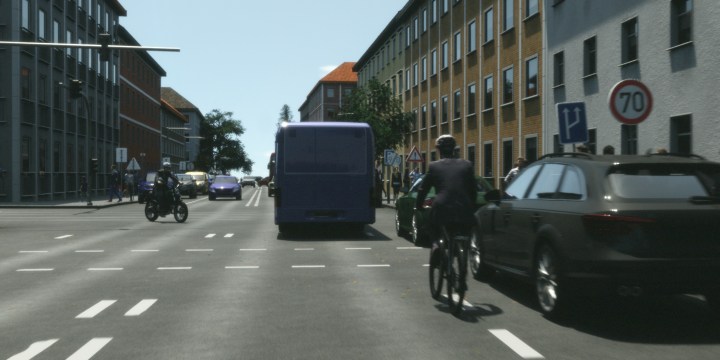}
\includegraphics[width=0.499\columnwidth]{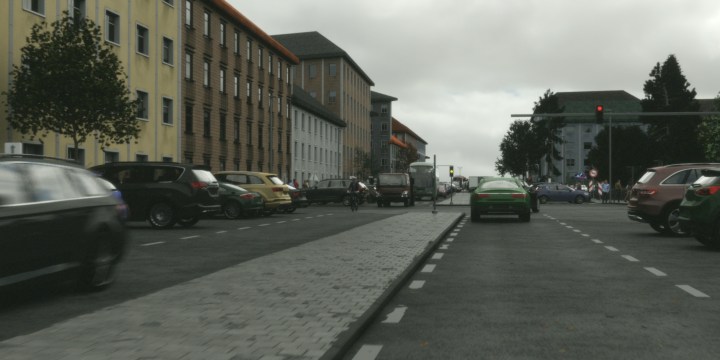}\includegraphics[width=0.499\columnwidth]{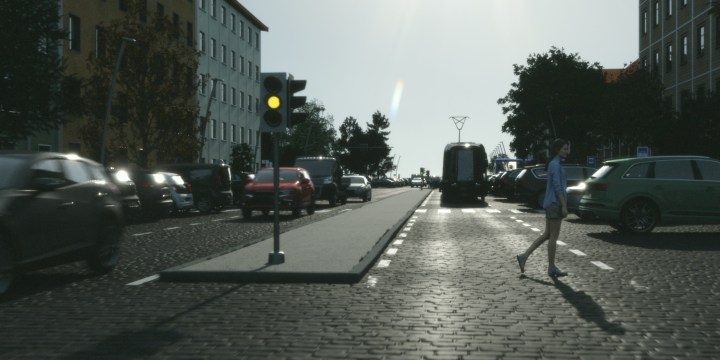}
\caption{Synscape's broad distribution over scenario parameters allows for selection of subsets along several scenario parameters at once. Left column: overcast, right column: sunny. Top to bottom shows increasing number of cars.\label{fig:ranges}}
\end{figure}

\begin{figure*}
\includegraphics[width=0.33\textwidth]{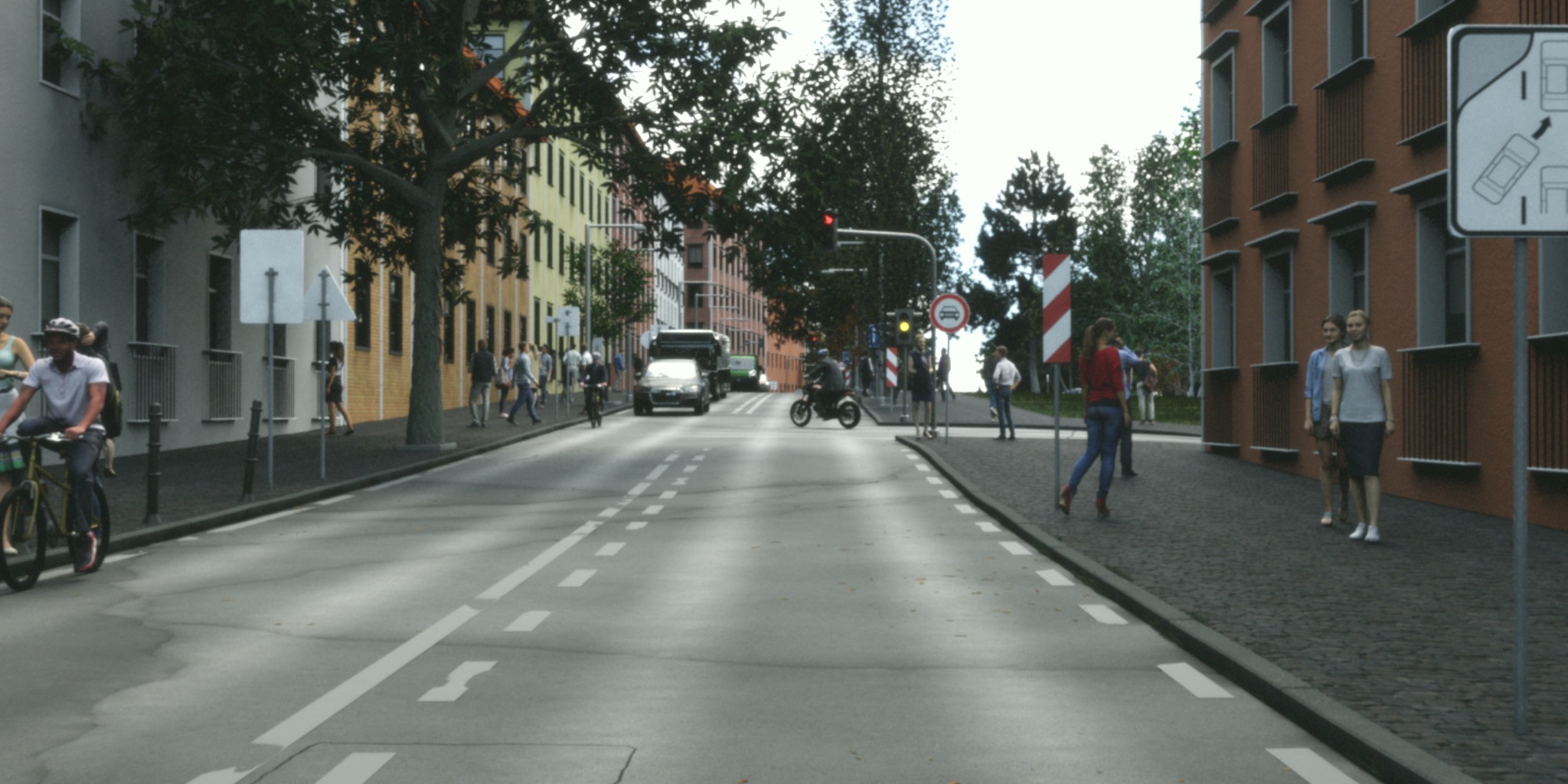}
\includegraphics[width=0.33\textwidth]{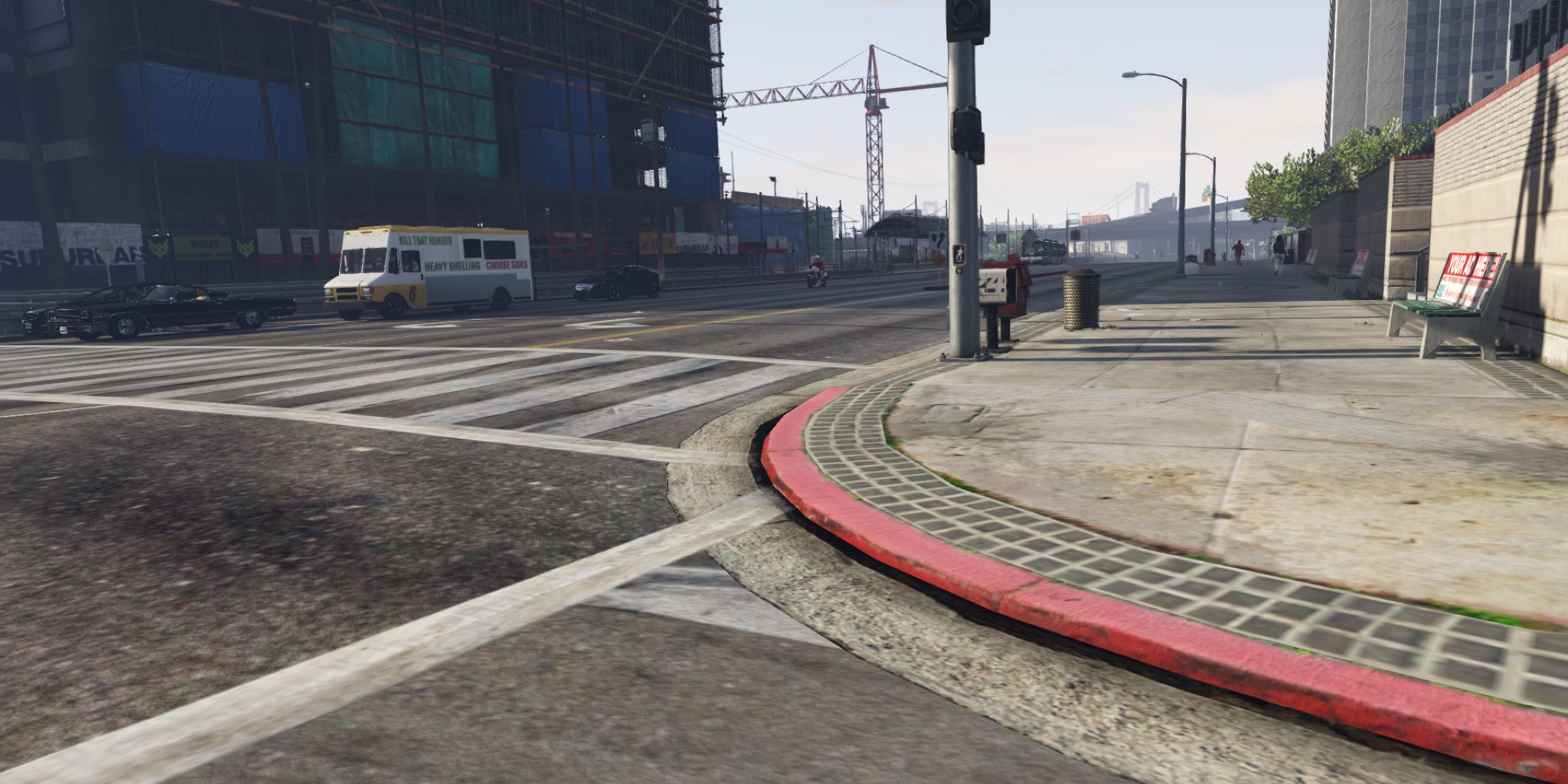}
\includegraphics[width=0.33\textwidth]{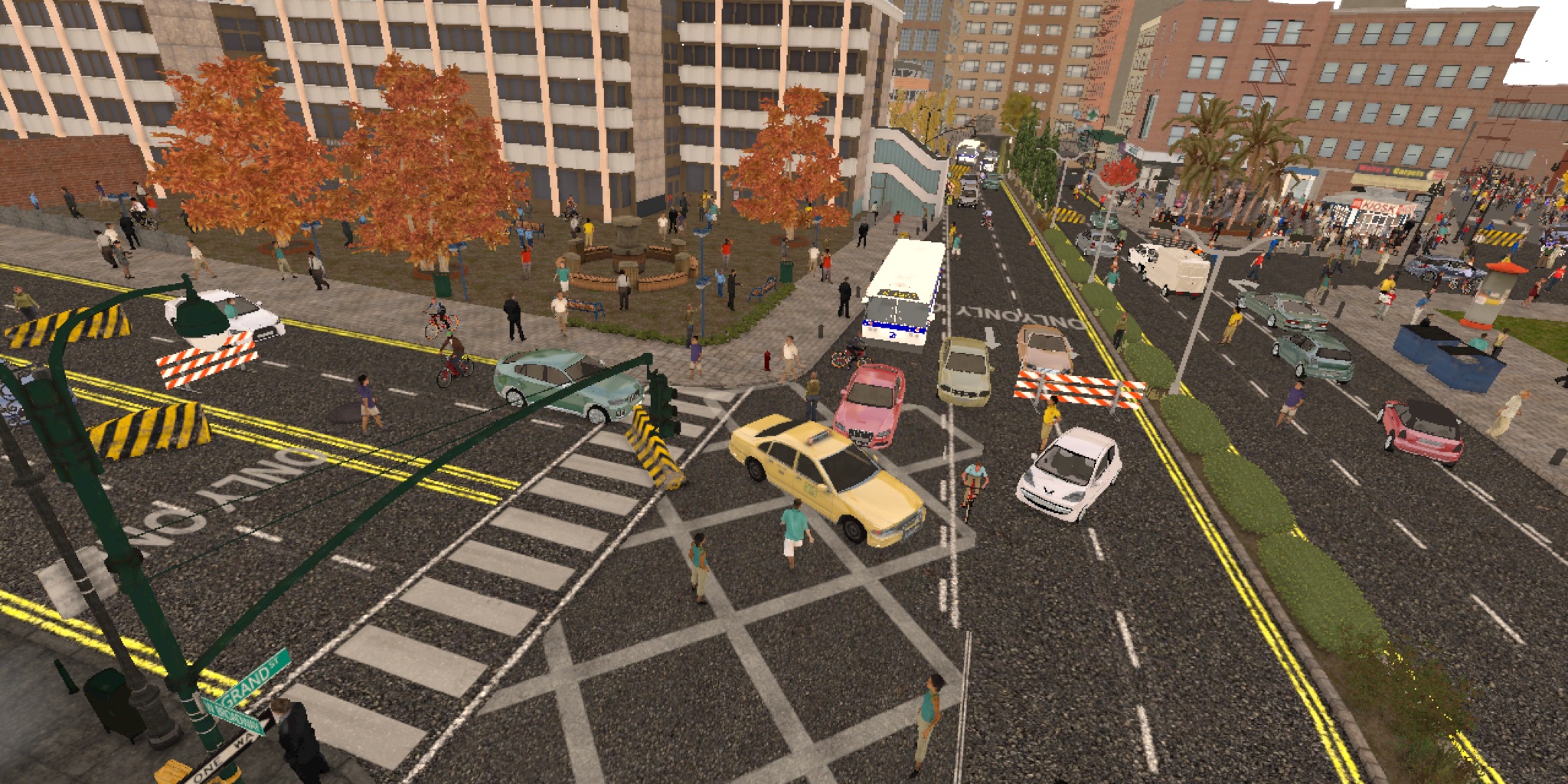}
\includegraphics[width=0.33\textwidth]{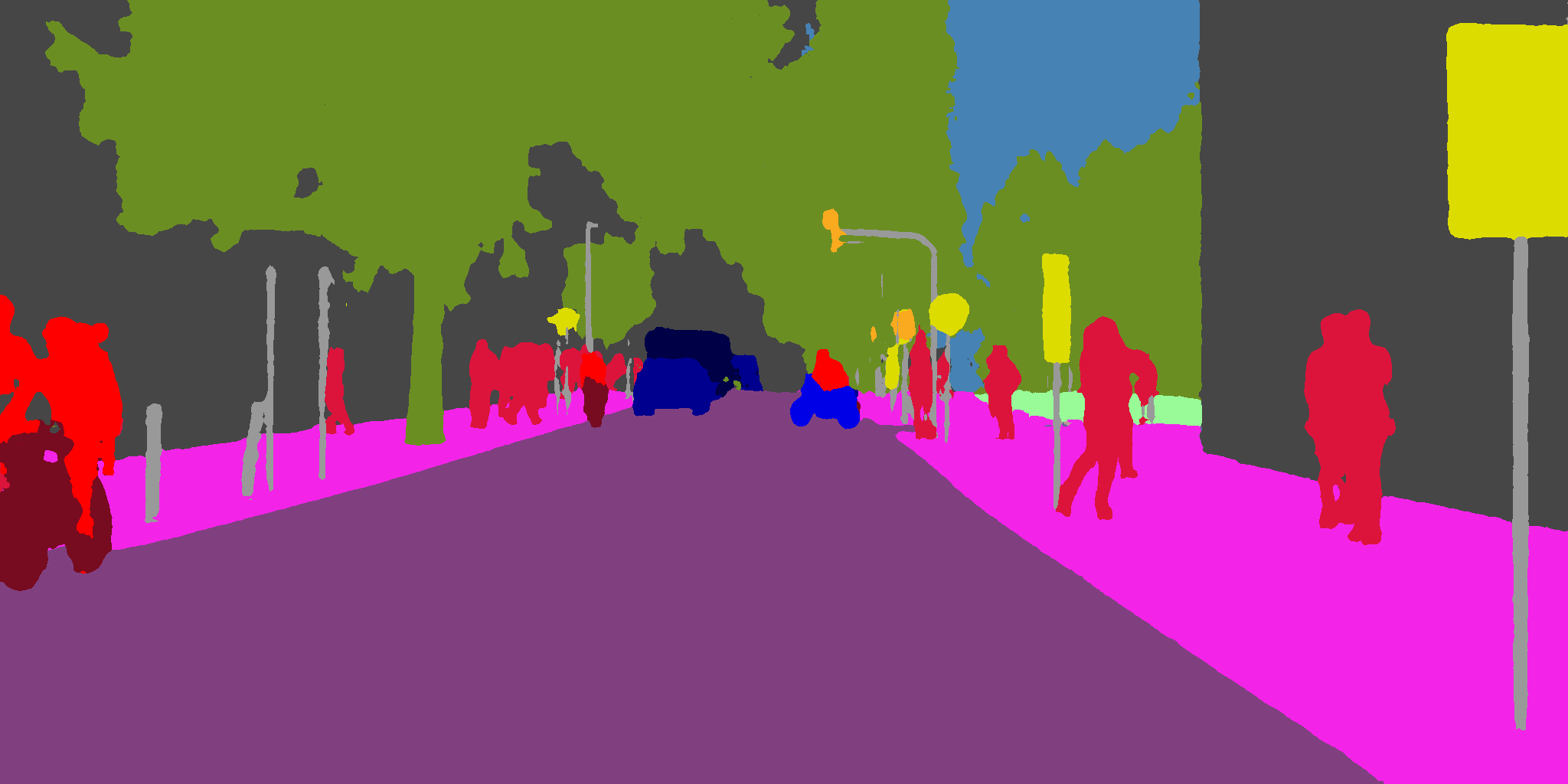}
\includegraphics[width=0.33\textwidth]{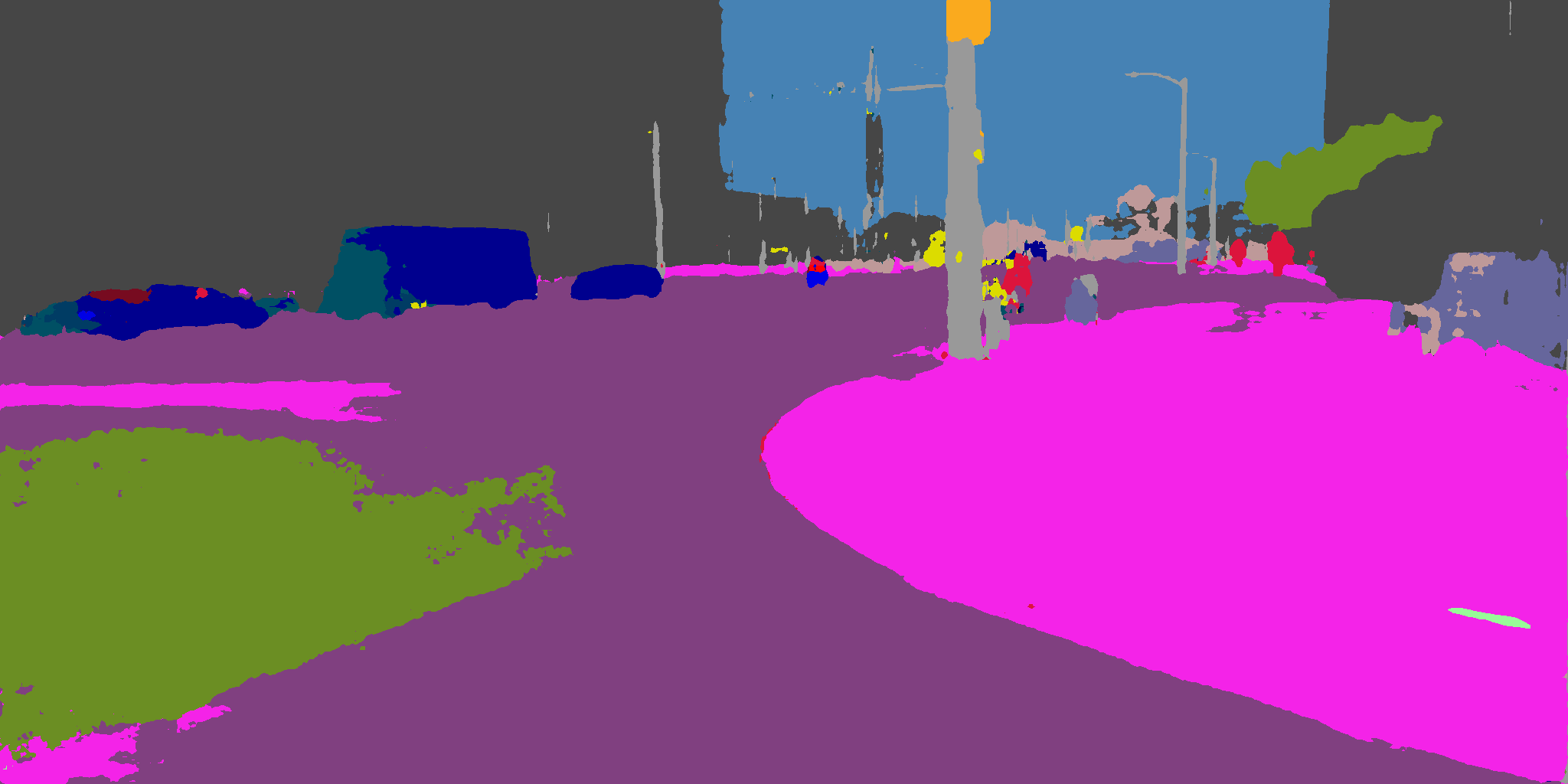}
\includegraphics[width=0.33\textwidth]{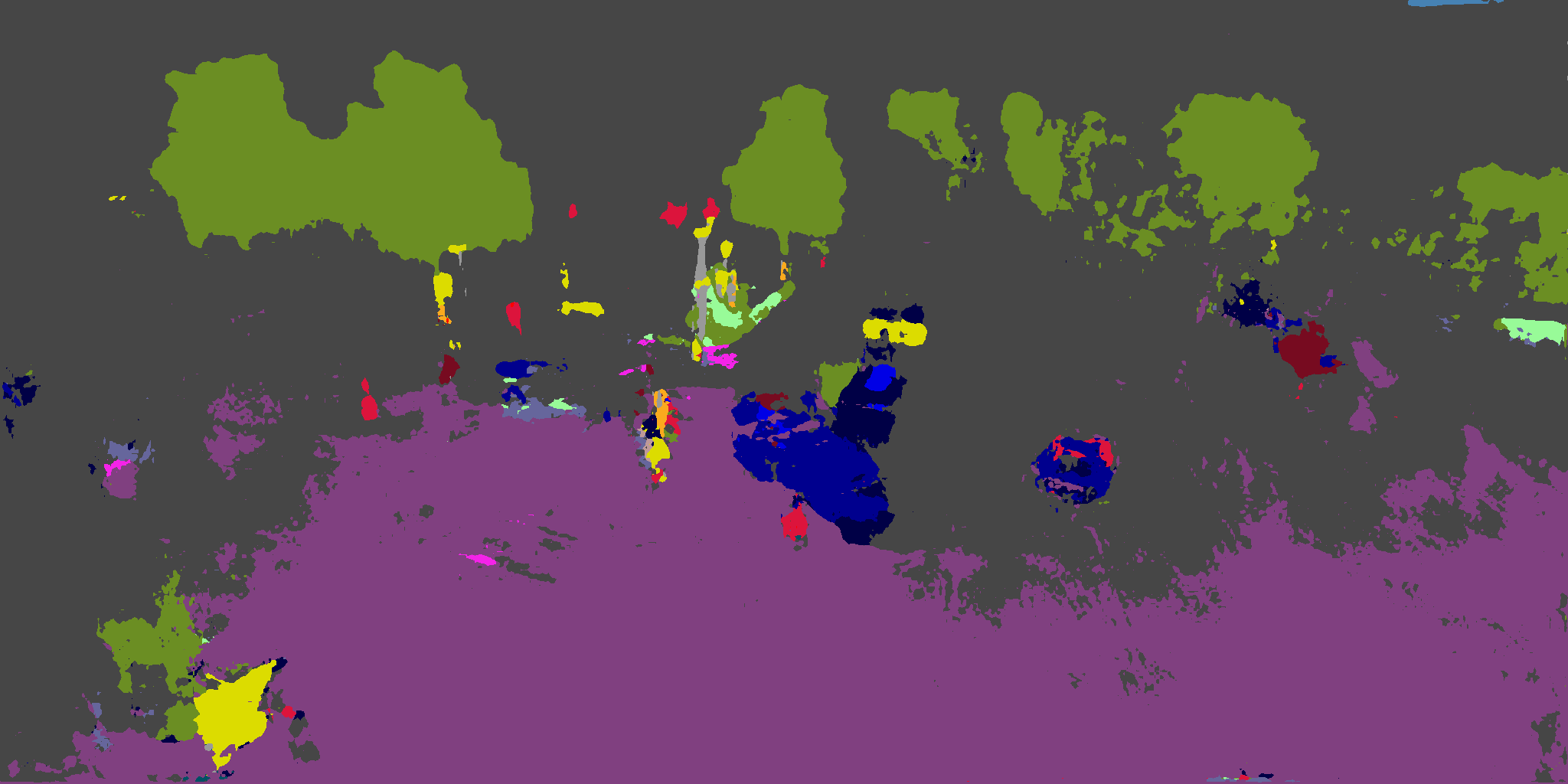}
\caption{Top row: Input images. Bottom row: predictions made by original authors' reference model of DeepLab v3+, trained on Cityscapes. Left to right: Synscapes, Richter (GTA) and Synthia. Table~\ref{table:validation_results} shows corresponding results.\label{fig:validation-images}}
\end{figure*}

\input{table-validation-short}

\noindent Synscapes was constructed in such a way that each scenario parameter is varied independently, providing a broad distribution across all dimensions of variation. In particular, care was taken to ensure that all scenario parameters are de-correlated. For example, if we wanted to study the difference between images near sunrise versus those taken with the sun at zenith, we still want the broadest possible distribution across all other scenario parameters. By using 25,000 unique scene variations for the 25,000 images we avoid unwanted correlations and make possible the sort of analysis studied in Section~\ref{sec:analysis}. Figure~\ref{fig:ranges} shows selected images across both of the \texttt{sky\_contrast} and \texttt{num\_cars} dimensions.

\subsection{Visualization scripts}

\noindent In order to visualize the instance metadata and view the dataset images sorted by scenario parameters, a set of scripts are provided\footnote{\href{https://github.com/7dlabs/synscapes-utils}{https://github.com/7dlabs/synscapes-utils}}. These also serve as a reference for how to extract and utilize the metadata, for example the projection of 2D and 3D bounding boxes into image space. 

\begin{itemize}
\item \texttt{view-synscapes.py} allows viewing of the dataset in sorted order according to any metadata parameter, e.g. according to sun height, or the number of visible cars. 
\item \texttt{visualize-synscape-metadata.py} can visualize the \texttt{class} and \texttt{instance} images as overlays on the RGB images, and also display 2D and 3D bounding boxes along with their respective class types. For clearer visualization, instances may be culled based on their occlusion level.
\end{itemize}


\section{Synthetic data in testing and validation}
\label{sec:validation}

\begin{figure*}
\centering
\includegraphics[width=0.495\textwidth]{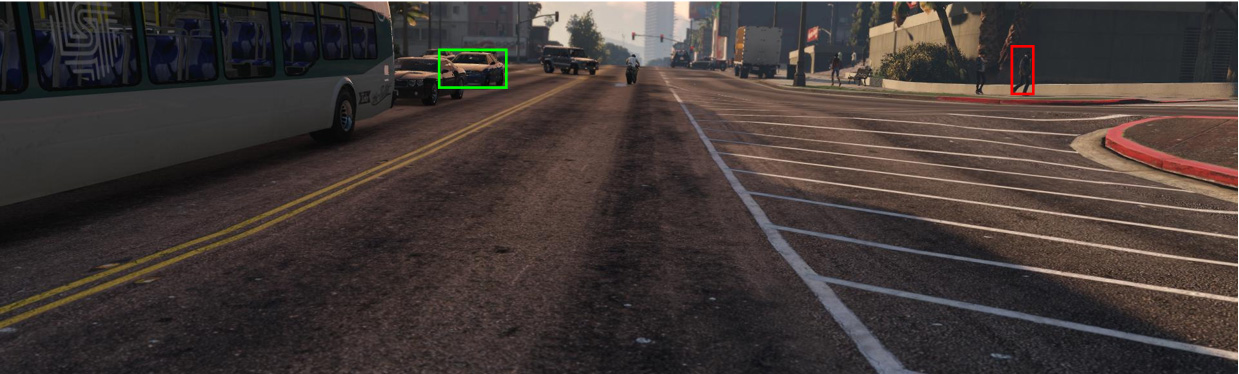} \includegraphics[width=0.495\textwidth]{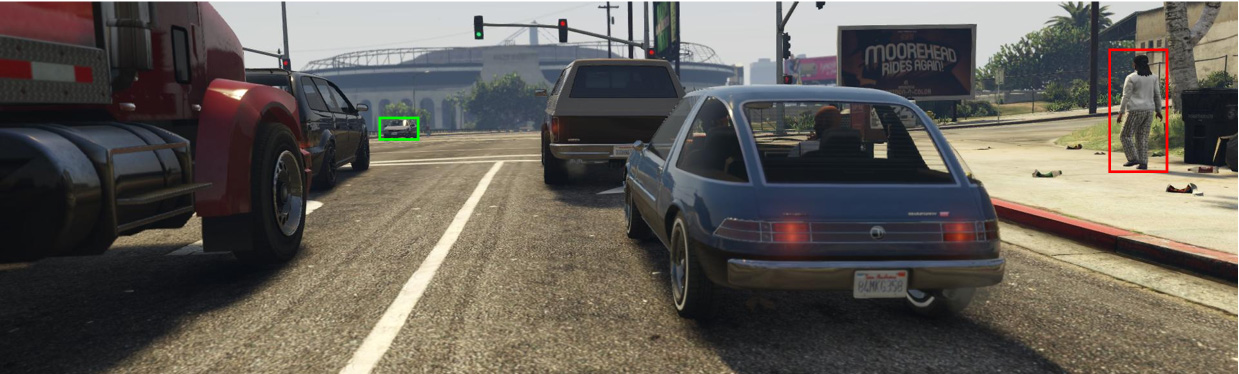}
\includegraphics[width=0.495\textwidth]{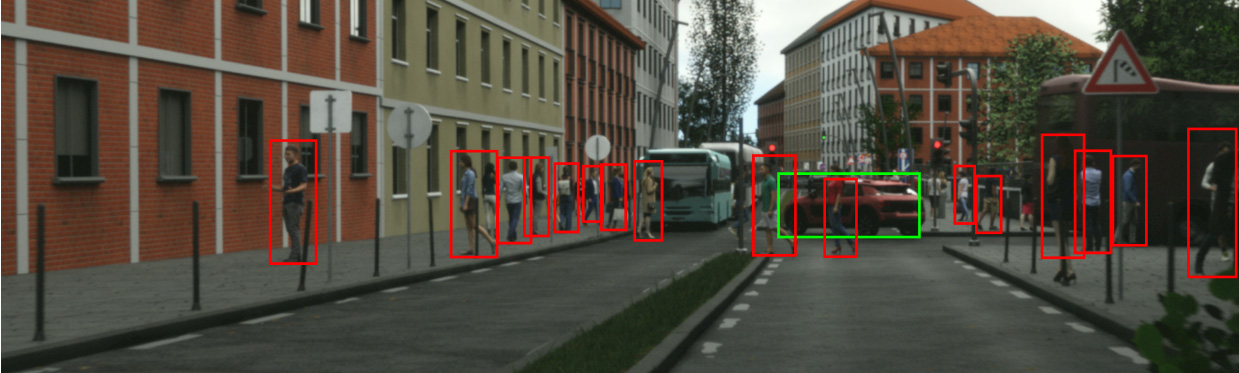} \includegraphics[width=0.495\textwidth]{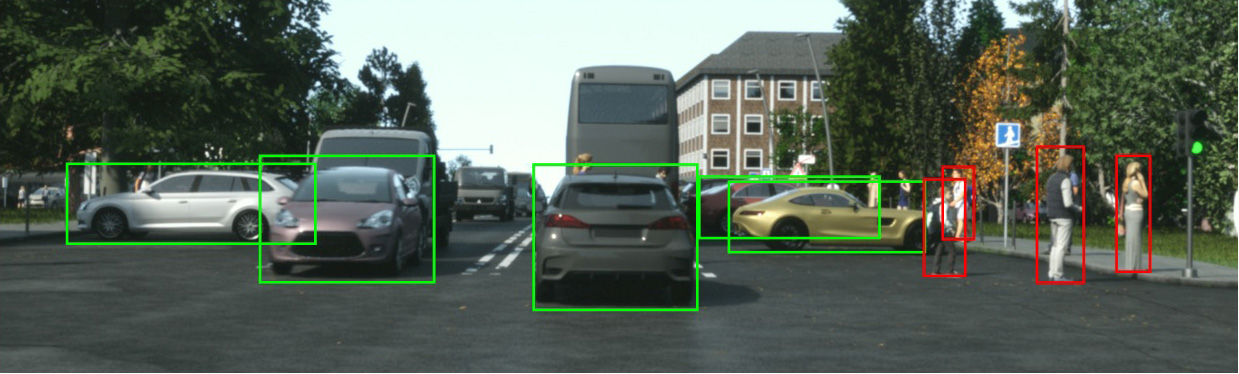}
\caption{Examples from object detection validation using Faster R-CNN trained on KITTI and evaluated on images from the GTA test set (top) and Synscapes test set (bottom). \label{fig:det_validation_images}}
\end{figure*}

\noindent One of the most common uses of synthetic data is in simulation. Whereas simulation of a planner can be done on a symbolic level, simulation of a perception system requires generation of sensor data that is both in the same format (quantitatively equivalent) and also of the same character (qualitatively equivalent) as real-world inputs. Simulating the data stream is generally straightforward, but simulation of the sensor's behavior as it reacts to (and potentially interacts with) its surroundings, either e.g. optically or electromagnetically, is both complicated and difficult to achieve. Still, it is a crucial puzzle to solve. If the data created by a virtual sensor doesn't correspond well to what its real-world counterpart would output, the simulation is not representative. For simulation of perception systems, this is a significant problem as it makes it difficult or impossible to determine whether e.g. the failure to detect a pedestrian in the simulation is due to domain shift (and that the identical real-world situation would be handled correctly), or whether there is a true deficiency in the model. In the end, this raises questions as to whether, and to what degree, sensor simulations can be trusted.

In order to quantify this effect we use publicly available networks that have been pre-trained on real-world datasets, and we then run inference on synthetic datasets. 
Although one may expect that synthetic datasets are less complex than real-world data and therefor easier to predict, the domain shift currently seems to obscure most such effects, with consistently lower scores on the synthetic datasets than the organic, real world counterparts that the models are trained on. Still, the effect likely plays a role, and finding methods that decouple domain shift from the relative difficulty of a given image would be a worthwhile research topic.

\subsection{Semantic segmentation}

\noindent We use the FRRN \cite{pohlen2017FRRN} and DeepLab v3+ \cite{deeplab3plus} architectures to evaluate performance on the semantic segmentation task. For each architecture we use the Cityscapes pre-trained models provided by the original authors and perform inference on the validation set of each synthetic dataset, along with Cityscapes itself for reference. 

As shown in Table~\ref{table:validation_results}, DeepLab achieves higher overall scores than FRRN (as expected), both overall and for individual classes. Furthermore, looking at the scores for each of the synthetic datasets, we see that they are consistently ordered for both network architectures. This suggests that their relative performances are a result of differences in the synthetic data rather than due to the network architecture itself. The Cityscapes-trained networks both achieve the best overall performance on Synscapes. This again suggests (but isn't by itself proof) that the domain shift is smaller for Synscapes than the other two datasets, and that visual realism has a significant impact on synthetic data's applicability as a testing and validation tool. Figure~\ref{fig:validation-images} shows examples of predictions for the DeepLab architecture on Synscapes, Richter and Synthia.

\subsection{Object detection}

\input{table-det-validation1}
\input{table-det-validation2}

\noindent To investigate the use of synthetic data as testing or validation for object detection, we use the Faster R-CNN architecture with Resnet101,~\cite{NIPS2015_5638} with pre-trained weights from the Google model zoo\footnote{\href{https://github.com/tensorflow/models/tree/master/research/object\_detection}{https://github.com/tensorflow/models/tree/master/research/object\_detection}} as a reference model. Detection is performed for the two KITTI classes \emph{car} and \emph{pedestrian} (pedestrian = person in Synscapes' labeling).

\begin{figure*}
\includegraphics[width=0.4985\textwidth]{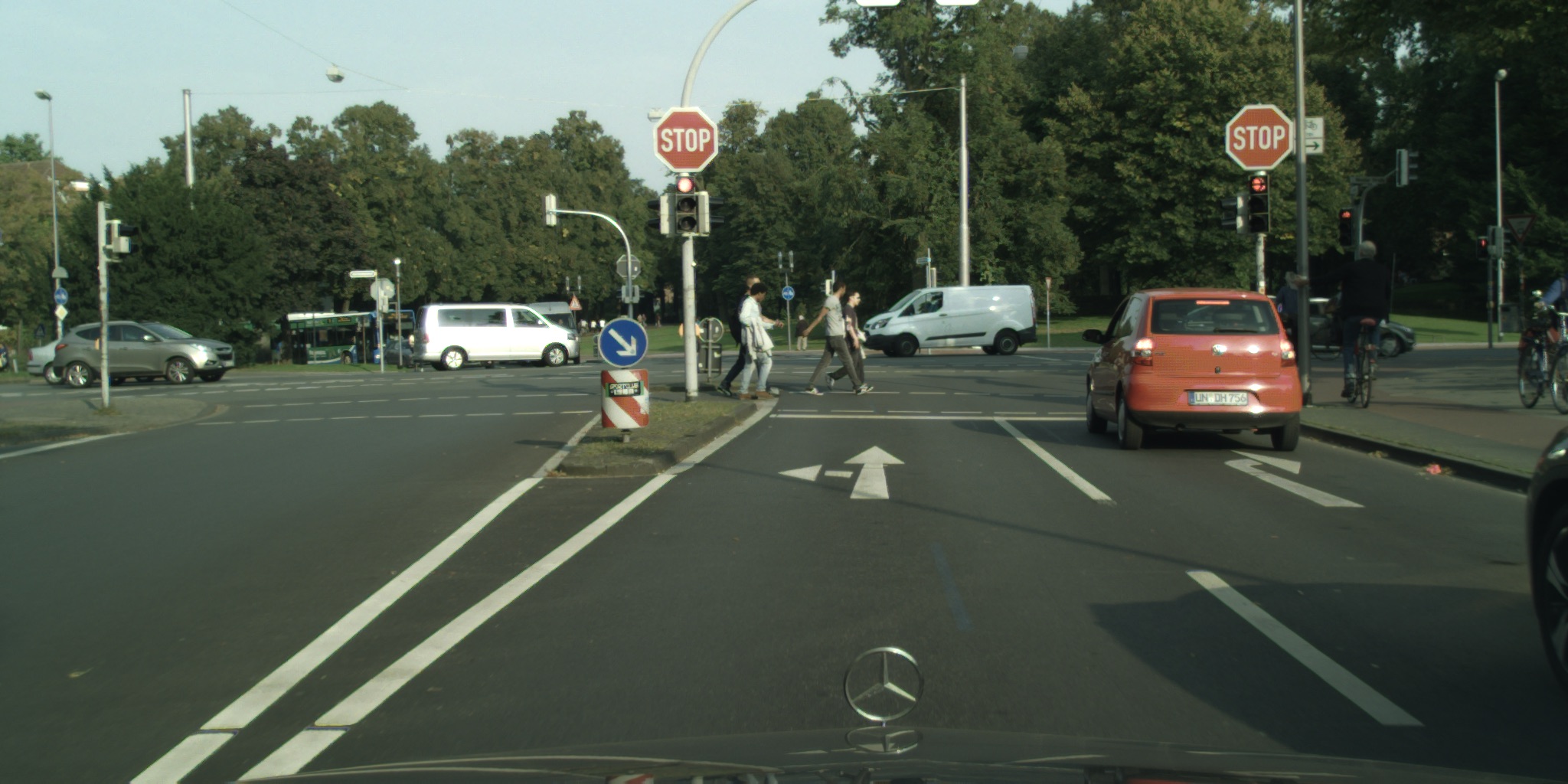}
\includegraphics[width=0.4985\textwidth]{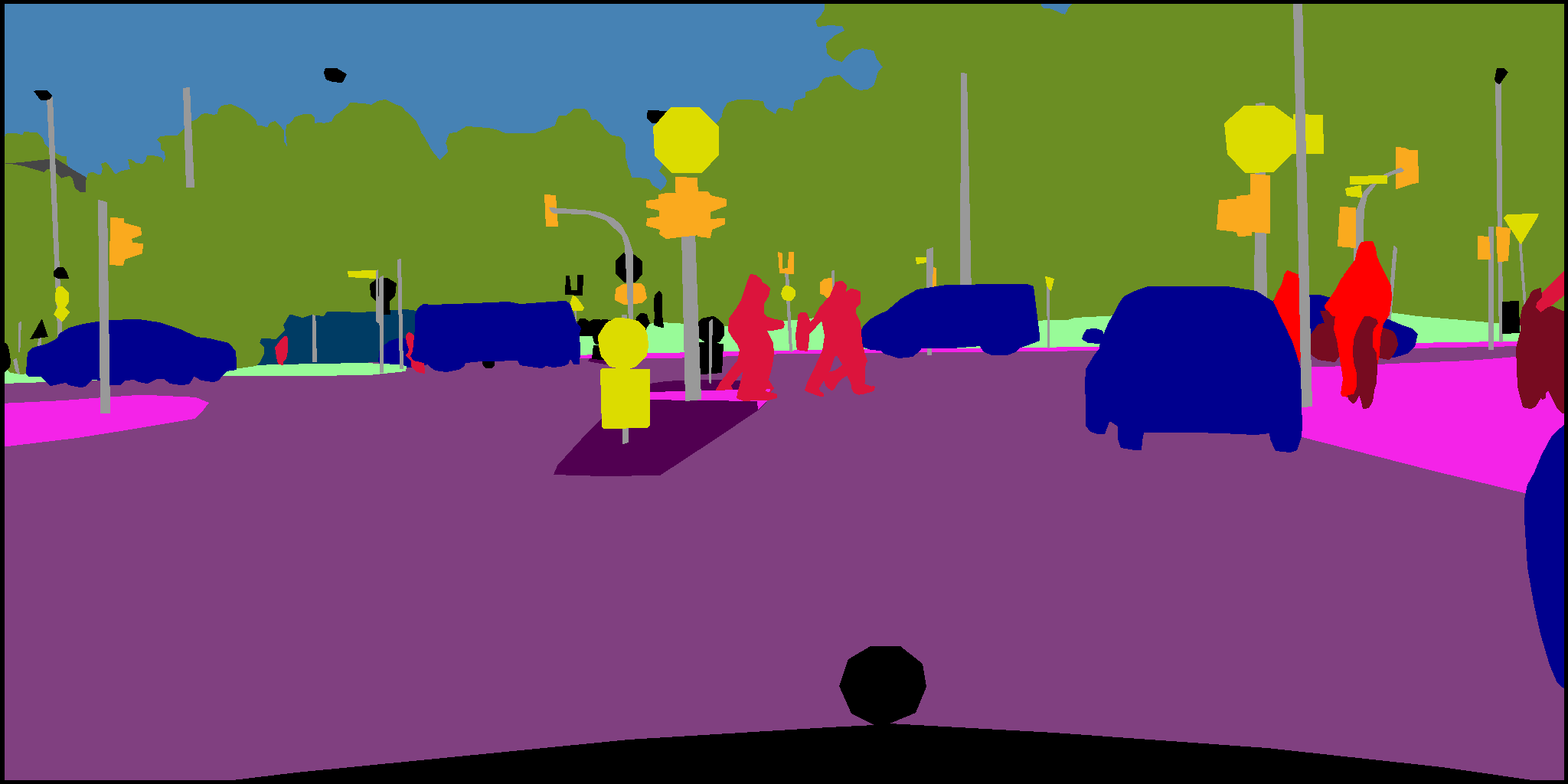}
\includegraphics[width=0.331\textwidth]{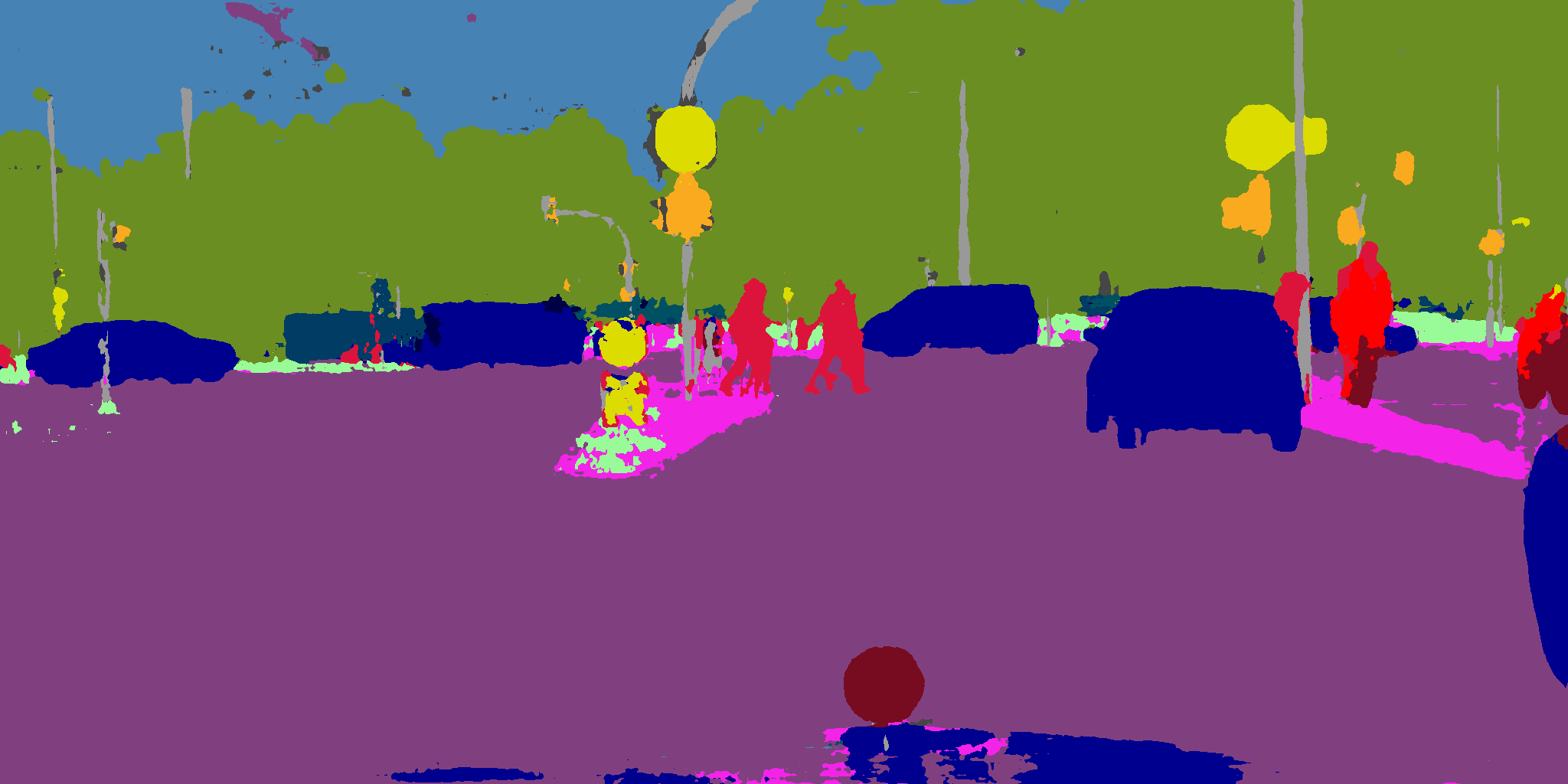}
\includegraphics[width=0.331\textwidth]{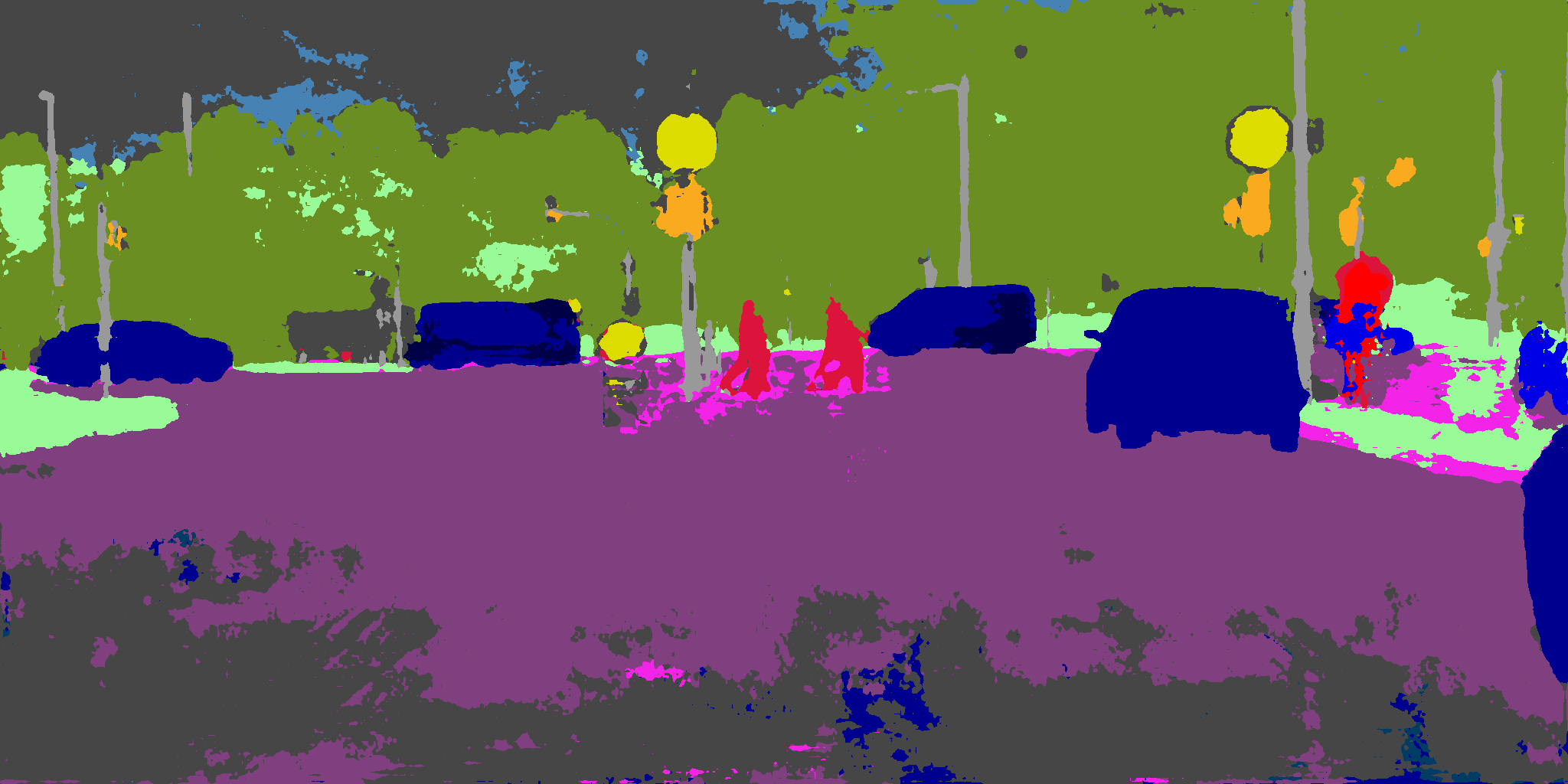}
\includegraphics[width=0.331\textwidth]{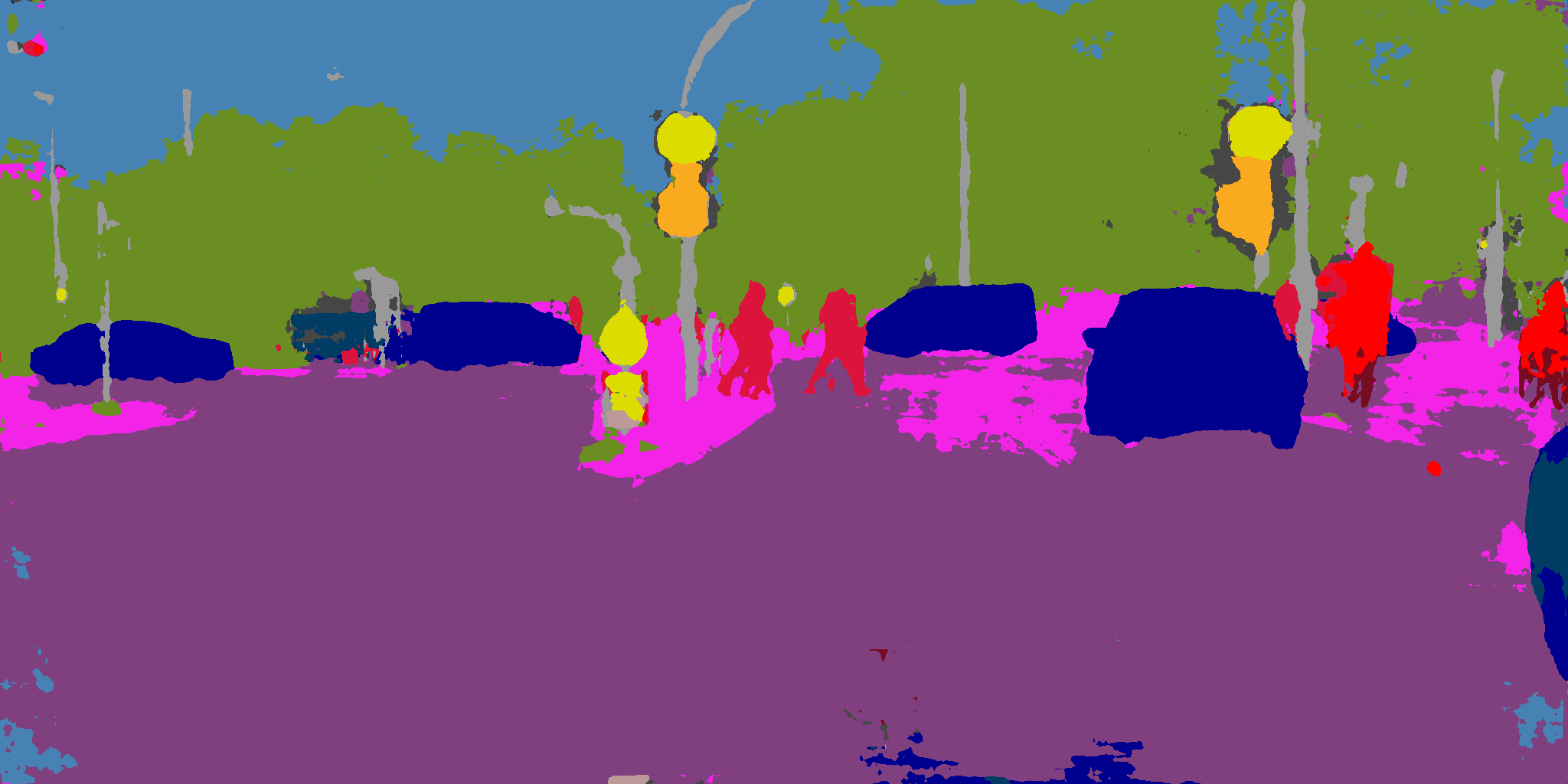}
\caption{Examples of predictions from DeepLab networks trained on synthetic data. Top row: Example image and ground truth annotations. Bottom row, left to right: Synscapes, Richter (GTA) and Synthia as training set. \label{fig:training-images}}
\end{figure*}

The results in Table~\ref{table:validation_det_1} show how object detection performs when trained on Synscapes in comparison to the Playing for Benchmarks dataset (denoted GTA),~\cite{Richter_2017}. GTA represents the current state-of-the-art for game engine-based approaches. For both datasets, the test set used consists of 1,000 images. For the GTA dataset, we maximized scenario/feature coverage by randomly selecting 1,000 images from the over 130,000 images in the database. The higher detection scores on Synscapes (mAP = 0.206) compared to GTA (mAP = 0.061) can most likely be attributed to the more accurate sensor simulation and better designed feature variation. The individual class scores for car and pedestrian displayed in Table~\ref{table:validation_det2}. Figure~\ref{fig:det_validation_images} displays example images with detections from the two test sets.

The more accurate imaging sensor simulation and variation in the Synscape dataset make the images more suitable for testing and analysis. As mentioned in Section~\ref{sec:overview}, Synscapes was designed with the Cityscapes dataset in mind, and there is a significant domain shift between KITTI and Cityscapes, which e.g.~can be seen in the differences in dynamic range and local contrast in the images, as well as in the density of car instances. By fine tuning the KITTI trained reference model, the results improve drastically. The results show that pre-training on KITTI and fine-tuning on Synscapes (KITTI + Synscapes) yields an increase in performance compared to training and validating only on Synscapes. While this is familiar for synthetic-to-real transfer learning, the reverse is not well explored.

\section{Synthetic data for training}\label{sec:training}

\input{table-training}

\input{table-validation}

\noindent Besides their use as testing and validation platforms, simulators often also promise the possibility of acting as a source for training data. In an ideal world, synthetic data would be superior to organic data, as detailed and accurate annotations could be generated at large volumes, and an arbitrarily large set of scenarios could be tested in parallel. In reality, however, there are several factors limiting this. Domain shift is perhaps an even larger problem in training than in testing, as is the challenge of creating virtual worlds with enough variation to provide meaningful learning material at scale.  

In order to analyze how different datasets perform as sources of training material, we train multiple state-of-the-art network architectures for semantic segmentation and object detection and compare how they perform when evaluated on organic, real-world data. In order to avoid chasing performance numbers as a goal in itself, we use the same hyperparameters for each training session, with the aim to give an uncolored view into each datasets' strengths and weaknesses.

\subsection{Semantic segmentation}

\noindent We train the same two networks as in Section~\ref{sec:validation} on Synscapes as well as the Richter and Synthia datasets. FRRN was trained with the default augmentation settings provided with the reference implementation, and was run for 100,000 iterations at a learning rate of $10^{-3}$ with a batch size of 3 on a single GPU. For fine tuning, the best synthetic-only training iteration was chosen, and a second set of 100,000 iterations was then run on Cityscapes. For DeepLab, training was run on 4 GPUs for 100,000 iterations with a crop size of 513, learning rate $10^{-2}$ and batch size 20.  

Table~\ref{table:finetuning_results} shows the per-class and mean IoU scores for the Cityscapes validation set. As with the validation tests in Section~\ref{sec:validation}, the results from both network architectures are consistent, with Synscapes producing the highest overall and per-class performance for all classes but one with both FRRN and DeepLab. If the higher validation score had been a function of Synscapes being "easier" than the other datasets, we would have expected the inverse effect during training. Instead, the opposite seems to hold.
The fine tuning results show a tighter race, but when looking at the relative improvement from the baseline (76.56\%) Synscapes provides more than twice the gain compared to both Richter and Synscapes.
Figure~\ref{fig:training-images} shows predictions on Cityscapes for DeepLab networks trained on the three datasets.

Finally, we study the performance for self-validation on the DeepLab networks. In Table~\ref{table:cross_validation_results} we note that Synscapes achieves a significantly higher score when evaluated on itself (87\% versus 63\% and 57\% for Richter and Synthia, respectively). 
Several factors are likely involved: Synthia and Richter's reduced geometric and rendering complexity likely results in a smaller feature space. Polygonal edges are clearly visible, and it is conceivable that the network confuses different classes due to these inadvertent characteristics. The distribution of objects and class imbalance is also likely at play. In contrast, Synscapes' higher realism avoids the first problem, and its broad distribution of scenario parameters improves on the second. 
As a measurement of whether all classes in a dataset are equally well recognized, we compute the standard deviation for classes that are present in the dataset and find $\sigma=8.25$ for Synscapes, and 17.51 and 24.55 for Richter and Synthia. 

\subsection{Object Detection}

\input{table-det-training1}
\input{table-det-training2}
\input{table-det-training3}

\noindent We train two architectures for object detection, Faster R-CNN with Resnet101,~\cite{NIPS2015_5638}, using the Tensorflow object detection API from Google\footnote{\href{https://github.com/tensorflow/models/tree/master/research/object\_detection}{https://github.com/tensorflow/models/tree/master/research/object\_detection}}, and the KittiBox implementation\footnote{\href{https://github.com/MarvinTeichmann/KittiBox}{https://github.com/MarvinTeichmann/KittiBox}} of FastBox~\cite{DBLP:journals/corr/TeichmannWZCU16}. With Faster R-CNN, we evaluate the training performance using the KITTI benchmark dataset ~\cite{Geiger12} as baseline. 
Tables~\ref{table:training_det1} and~\ref{table:training_det2} show  results from training on the two classes \emph{car} and \emph{pedestrian}, evaluated on test sets consisting if 500 images extracted from  both the KITTI data set and Synscapes respectively. The first column indicates which data set was used during training and/or fine tuning, and the second which test set was used.
For all training and fine tuning we used 300k iterations and a learning rate of $10^{-2}$ for iterations up to 200k~and $10^{-3}$ for iterations over 200k. Although training on Synscapes alone performs poorly on KITTI due to the domain shift emanating from the fact that Synscapes was designed with the Cityscapes dataset in mind, the COCO results obtained when applying fine tuning shows a significant increase in performance as compared to the baseline in the first row. Interestingly, the fine tuned results both when training on Synscapes and validating on KITTI (Table~\ref{table:training_det1}), and when training on KITTI and validating on Synscapes (Table~\ref{table:validation_det_1}) not only bridges the domain gap in both directions, but also increase the performance with respect to both baselines (KITTI - KITTI and Synscapes - Synscapes).

Table~\ref{table:training-det3} shows an evaluation using the KittiBox implementation of FastBox~\cite{DBLP:journals/corr/TeichmannWZCU16}. Training was performed with a learning rate of $10^{-5}$ and 250k iterations for both training on Synscapes and fine tuning. Also in this case, the use of synthetic data improves the performance significantly.

\section{Analysis using Synscapes}
\label{sec:analysis}

\noindent The procedural approach used to generate the imagery makes Synscapes particularly suitable for algorithm and dataset analysis. The carefully controlled distribution of the generating parameters used in the scenario generation enables us to make cuts in, or \textit{bin} the resulting images and labels along the parameter dimensions. Another enabling factor is the wealth of metadata being generated for each rendered image.

Given the statistics in the feature variation in Synscapes, slicing, or binning along a dimension \emph{leaves even distributions along all other dimensions}. Figure~\ref{fig:ranges} illustrates how the weather conditions are binned into sunny and overcast skies, while the density of cars is quantized into bins ranging from a few to many cars in the image. Similarly, given the metadata one can for individual instances of objects/classes within the images slice along dimensions such as occlusion, heading, or distance from the ego-vehicle etc., which are useful in analyzing how an existing ML model reacts to varying inputs.

\subsection{Semantic segmentation}

\noindent In order to explore statistically how DeepLab's pre-trained Cityscapes model behaves, we run prediction on the Synscapes training set (24,000 images). By binning either the images or the individual instances in each image according to one or more of the scenario and instance parameters' values, we can average the performance of the network for all the images in a given bin, knowing that they represent a wide distribution of parameter values in all dimensions \textit{except the binned one}. 

\subsubsection{Effect of object orientation}

\begin{figure}
\centering
\includegraphics[width=1\columnwidth,trim={0 2cm 0 0cm}, clip]{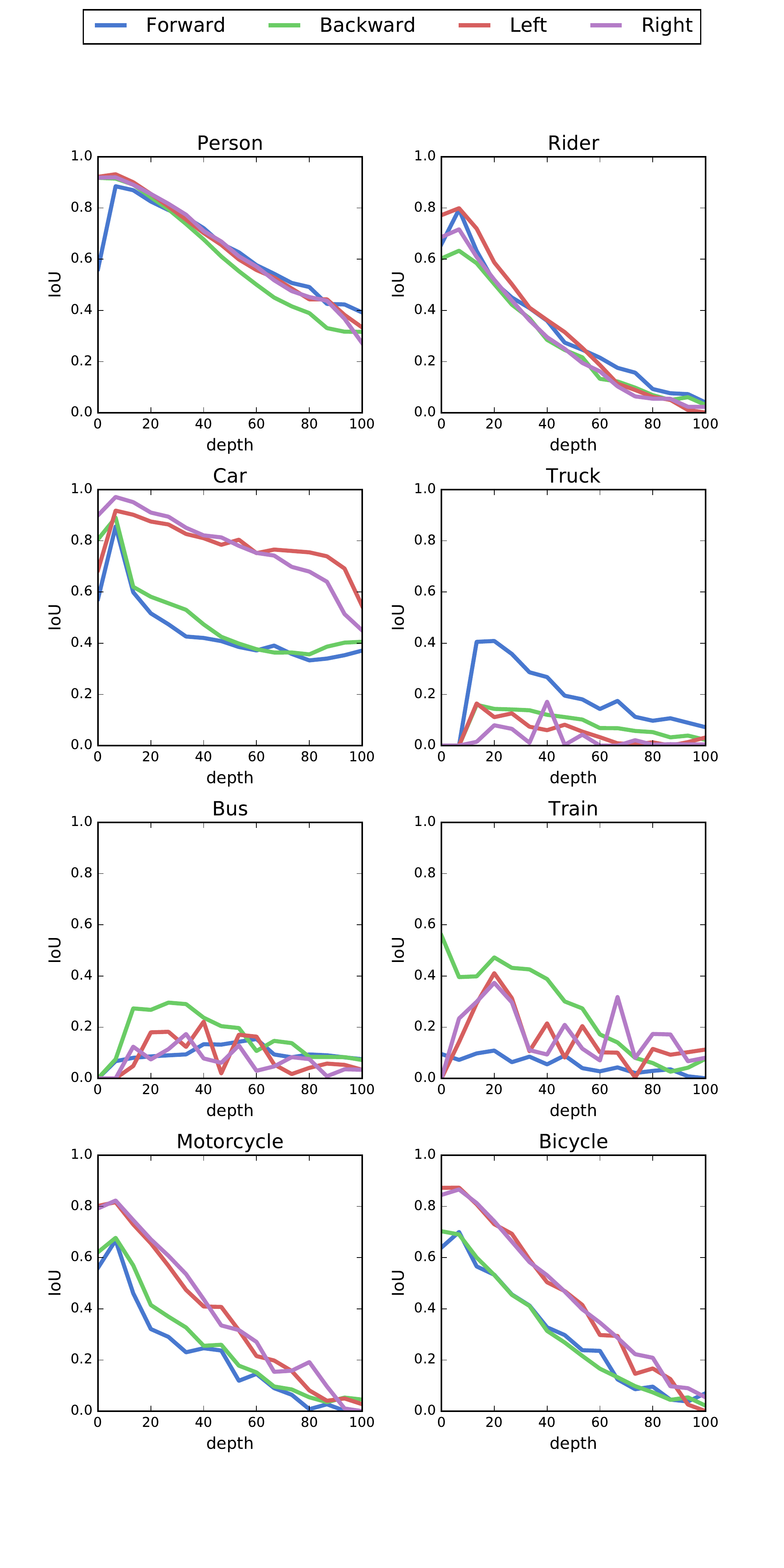}
\caption{Analyzing the effect of each instance's orientation. IoU score for the four basic orientations over depth. (Cityscapes-trained Deeplab v3+ on Synscapes.)\label{fig:orientation}}
\end{figure}

First, we use the oriented 3D bounding box to determine the relative orientation of each instance and study its effect on the segmentation performance. The pixels for each actor instance are categorized as belonging to one of the four cardinal directions (forward, backward, left or right relative to the ego vehicle), and a separate IoU score is computed for each subset. Additionally, we divide the predicted pixels of each direction category into 16 segments along depth to show how the performance also varies as a function of depth, as seen in Figure~\ref{fig:orientation}.

In this visualization, we see that the Person and Rider classes are well recognized independently of view direction, but for Car, Motorcycle and Bicycle the network performs worse for instances in the forward/backward directions than left/right.

Truck, Bus and Train all show differences between oncoming and same-side instances, which could be a factor both of bias in the training set distribution, but also due to the fact that e.g. a truck is easier to distinguish from a bus when seen from the rear, compared to when seen from the front.

\subsubsection{Effect of object occlusion}

\begin{figure}
\centering
\includegraphics[width=5cm]{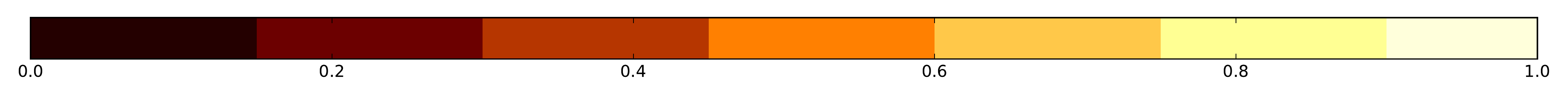}
\includegraphics[width=1\columnwidth,trim={0 2cm 0 1.5cm}, clip]{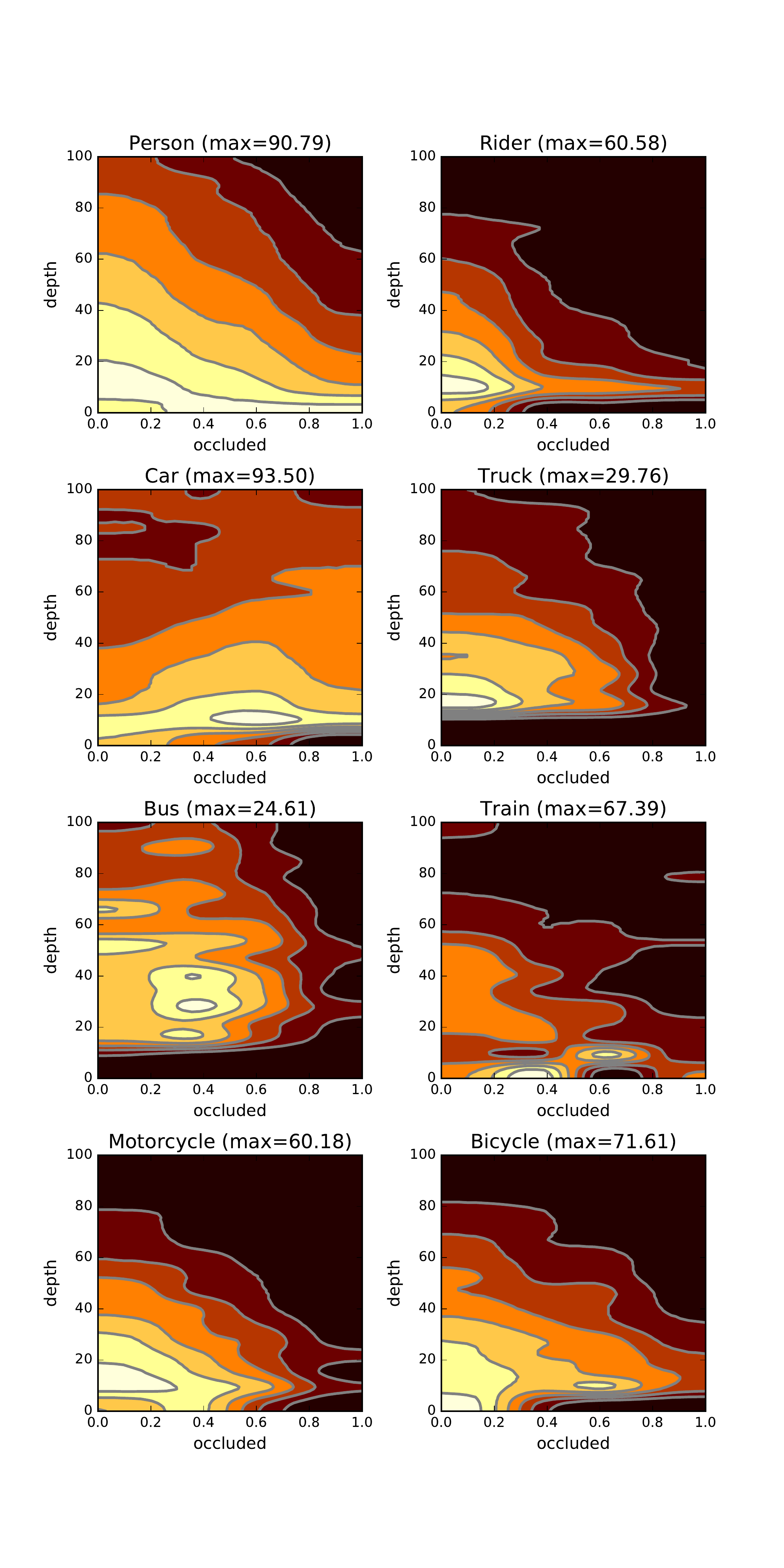}
\caption{Analyzing the effect of instance occlusion. The heat map shows relative IoU score across occlusion and depth. (Cityscapes-trained Deeplab v3+ on Synscapes.)\label{fig:occlusion}}
\end{figure}

\begin{figure*}
\centering
\includegraphics[width=\textwidth,trim={6cm 2cm 5cm 0},clip]{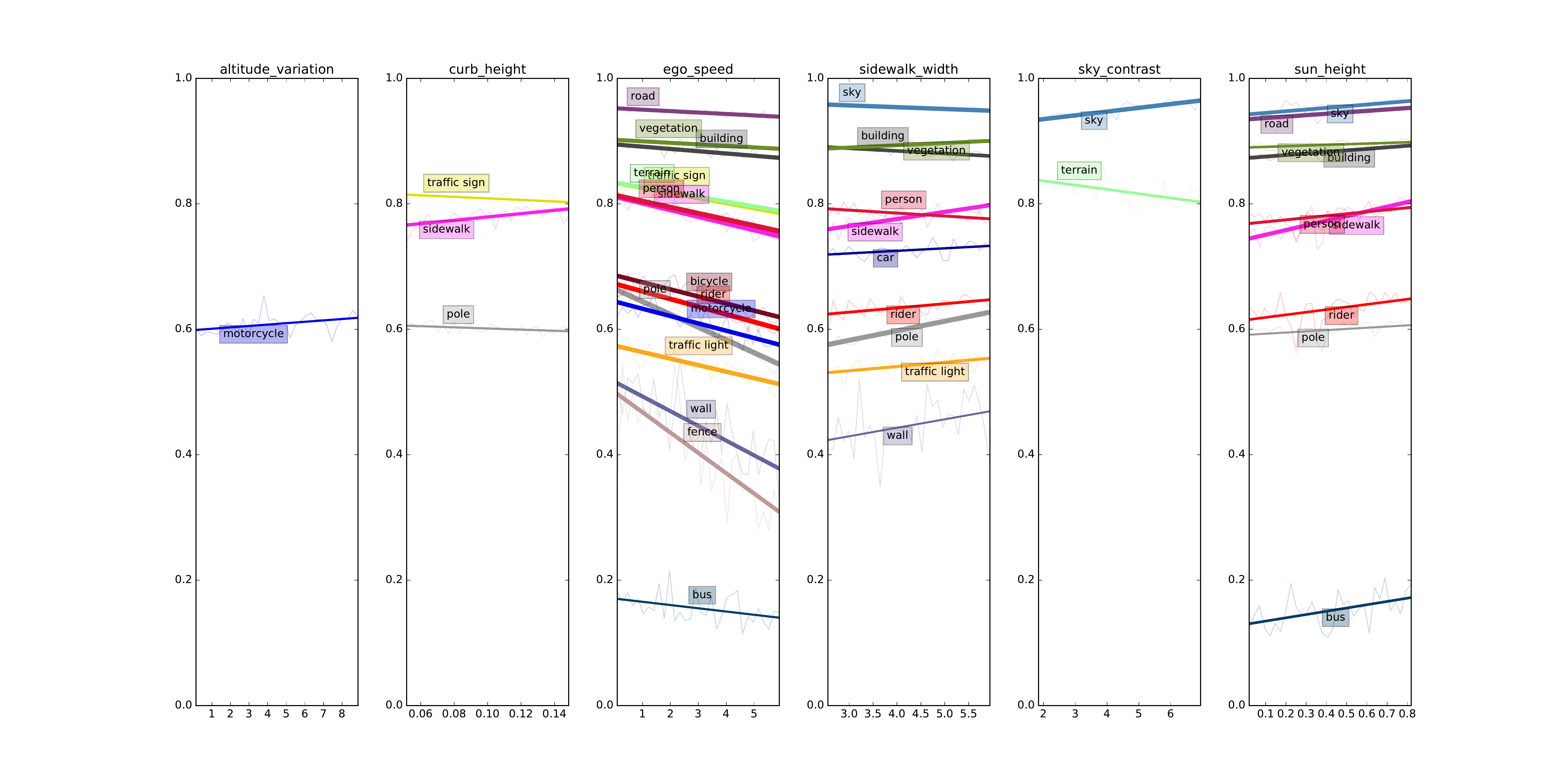}
\caption{Linear regression between meta-parameters and per-class IoU score for semantic segmentation. Line width indicates correlation coefficient, and only correlations with p-value 0.05 or less are included for clarity. Scores for the 64 individual subsets are overlaid to show variance. (Cityscapes-trained Deeplab v3+ on Synscapes.)\label{fig:correlation}}
\end{figure*}

Synscapes also provides a per-instance metric of the fraction of each object that is visible to camera. We first divide predicted pixels into four subsets according to the amount of occlusion: $[0.0,0.25]$, $[0.25,0.5]$, $[0.5,0.75]$, and $[0.75,1.0]$. Then, each subset is further divided according to depth. The IoU score for each subset is shown in Figure~\ref{fig:occlusion} as a heat map, which visualizes some inherent properties of the Cityscapes dataset.

We can see that Person and Rider score highest when unoccluded, whereas Car and Bus both have highest scores in areas with partial occlusion. This is likely due to differing exemplar balances in the training data, as Cityscapes is quite busy overall, with entirely unoccluded vehicles the exception rather than the rule. 

We also note that Rider, Motorcycle and Bicycle perform very similarly, as is expected given their correlated placement in the training data.

\subsubsection{Correlating performance to scenario parameters}

Clearly, a network will not achieve the same performance on all input images. Certain configurations are more likely to be analyzed correctly, others may be harder. This effect is due to many factors: whether a given feature is present in the dataset, whether the exemplars are observed in varied enough conditions, etc. Knowing which configurations are problematic is highly desirable but also difficult to achieve with organic data; annotating bounding boxes can be done precisely, but labeling abstract descriptions of an image, e.g. degree of fogginess or amount of wear and tear to a road surface, is difficult to do consistently. Finally, capturing sufficient amounts of data to make reliable analyses, especially for uncommon situations, is expensive.

In order to illustrate how Synscapes' detailed metadata and broad variation can be used to address this type of analysis, we explore whether there are correlations between the segmentation network's performance and the scenario parameters used to generate the synthetic data. For a given selection of parameters, we divide the predictions on the 24,000 images in the training set into 64 subsets, and compute an IoU score for each class. We then perform a standard linear regression, and record the correlation factor and p-value for each class/parameter pair. Pairs with a p-value over 0.05 are discarded, with Figure~\ref{fig:correlation} showing the resulting samples.

We find that motion blur (\texttt{ego\_speed}) and time of day (\texttt{sun\_height}) are the scenario parameters with the strongest effect on the network's predictive performance. As motion blur increases in an image, features are smeared, and although the result is easy for a human to interpret, the resulting features would appear very different to a convolution-based neural network. In particular, classes with strong vertical features, such as Pole, Wall and Fence, are most strongly affected. Although detection of the road surface degrades, it does so the least of the measured classes, likely because it is (in general) the closest to the ego vehicle and will therefor contain the most motion blurred features in the training set.

Similarly, as the sun approaches the horizon, overall contrast in the image is reduced. Because of auto-exposure, the image isn't necessarily darker, but without strong shadows features are generally more difficult to distinguish. We note the difference between overcast conditions (which illuminate the scene more evenly) from sunset conditions, which may have high contrast. Finally, we also see an expected correlation between the curb height and the score for Sidewalk, as the higher curb makes the edge more distinguishable.

\subsection{Object detection}

\noindent In order to do a similar study for object detection, we use a reference model of a Faster R-CNN network trained on KITTI and evaluate performance on 20,000 Synscapes images.

\subsubsection{Effect of object orientation}

\begin{figure}
\centering
\includegraphics[width=0.9\columnwidth]{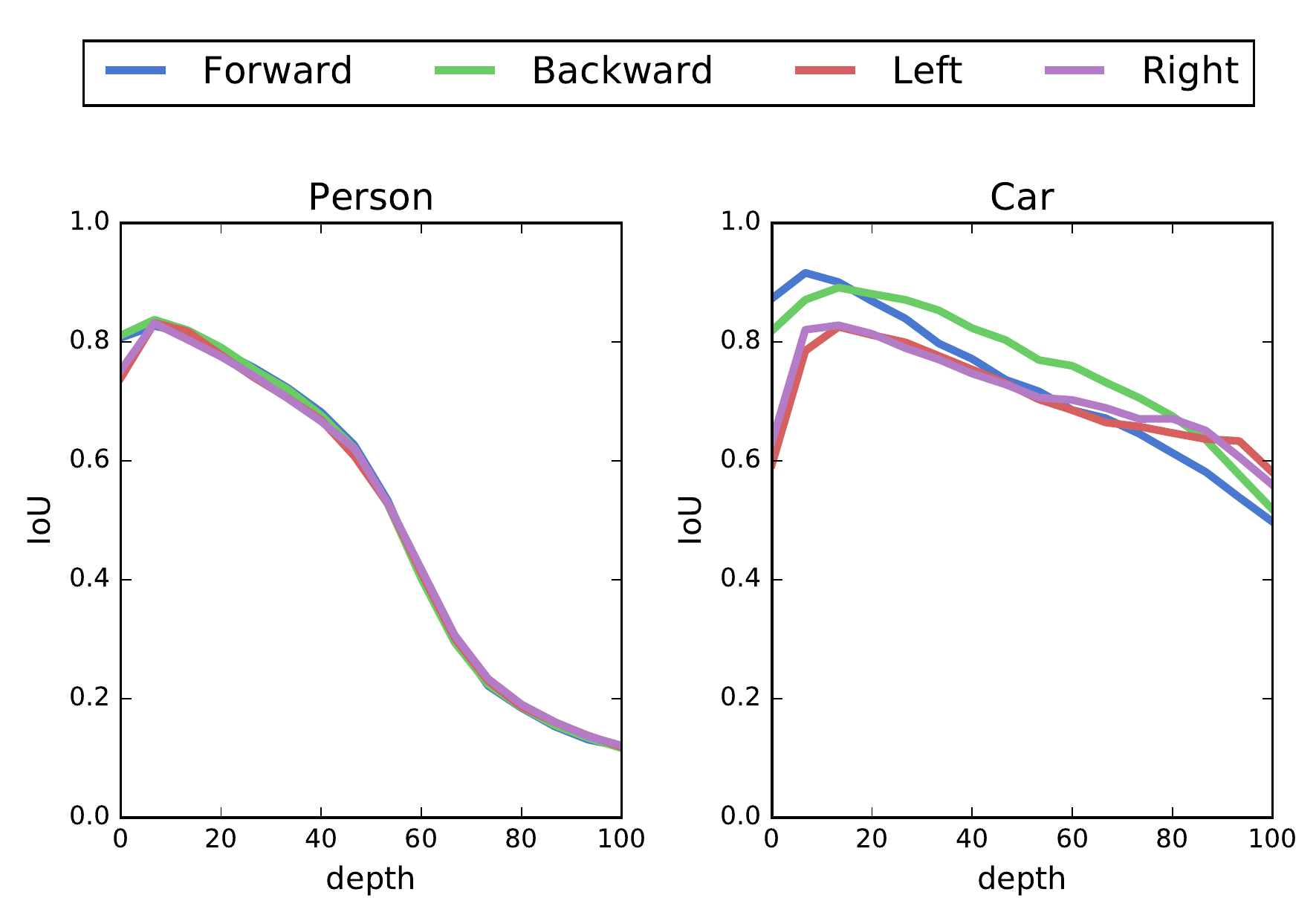}
\caption{Analyzing the effect of instance orientation. IoU score for the four basic orientations over depth. (Faster R-CNN KITTI-trained model evaluated on Synscapes.)\label{fig:det-orientation}}
\end{figure}

Compared to the Cityscapes-based DeepLab model, we see some similarities and also some differences. First of all, as shown in Figure~\ref{fig:det-orientation}, the Pedestrian class is detected consistently for all four directions, although there is a sharp drop-off in performance at a distance of around 50 meters, whereas the segmentation model's performance degraded more linearly with distance. 

Similarly to the segmentation model, the Car class has distinct directional dependence. However, where Cityscapes has significantly better performance on cars seen from the side, the difference for the detection network is more subtle, and performs best for cars facing the ego vehicle (heading backwards), followed by cars facing the same direction. Interestingly, beyond 80 meters, cars are detected somewhat more reliably when seen from the side.

\subsubsection{Effect of object occlusion}

\begin{figure}
\centering
\includegraphics[width=5cm]{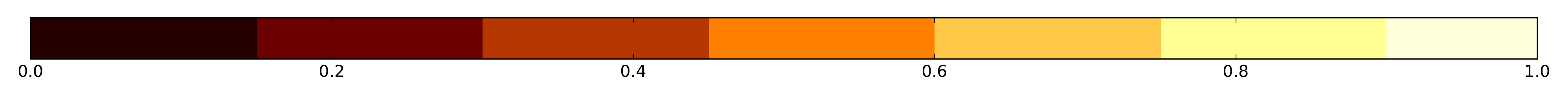}
\includegraphics[width=1\columnwidth]{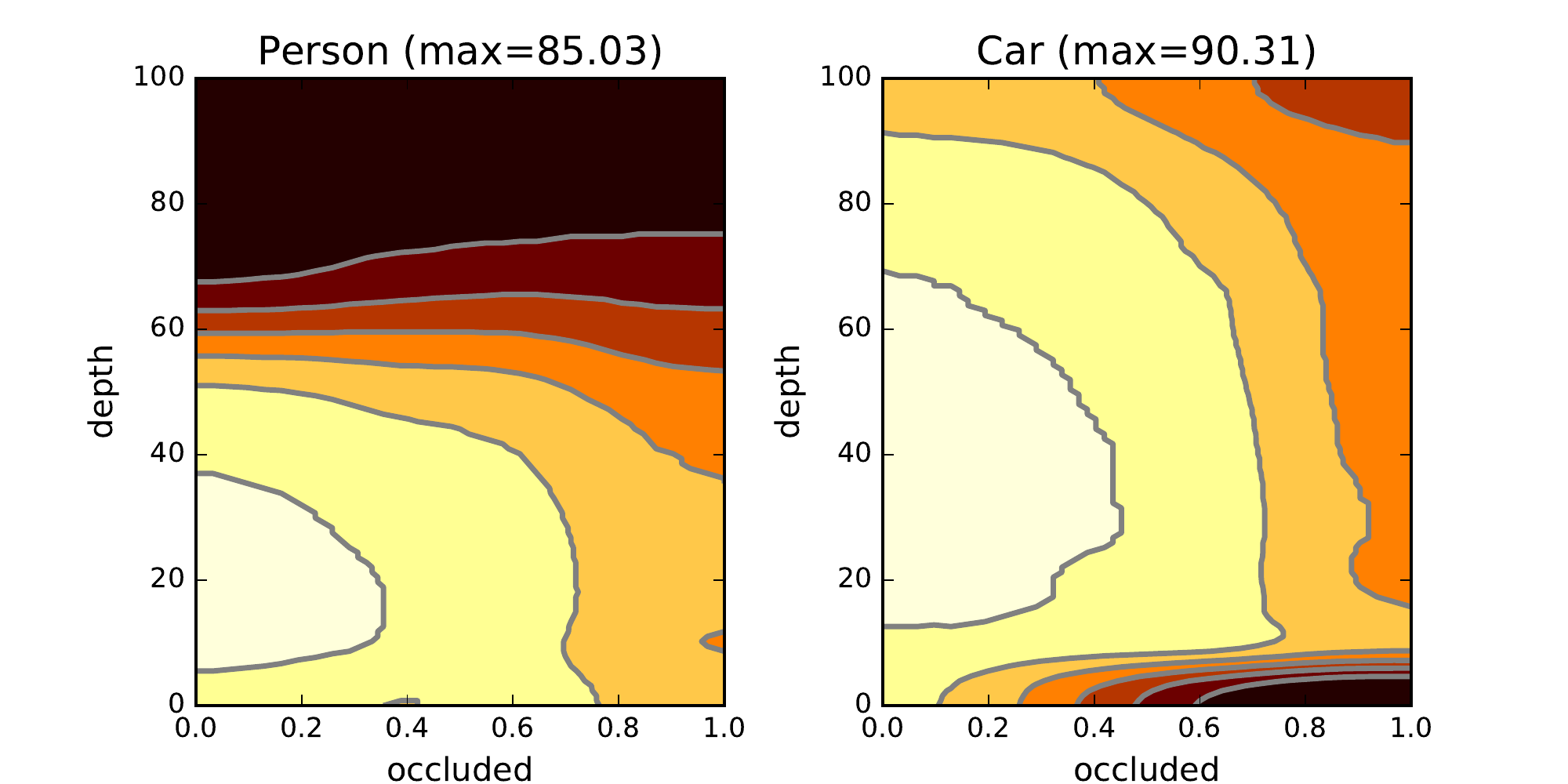}
\caption{Analyzing the effect of instance occlusion. The heat map shows relative IoU score across occlusion and depth. (Faster R-CNN KITTI-trained model evaluated on Synscapes.)\label{fig:det-occlusion}}
\end{figure}

\noindent Figure~\ref{fig:det-occlusion} shows a heat map of the Person and Car classes. The falloff in performance with distance is again clear, and we can see that e.g. an 80\% occluded person at 10 meters is as difficult to detect as an unoccluded person at more than 50 meters. For distances up to 40 meters, performance on the Person class is strongly affected by occlusion, but beyond this range occluded instances are detected as well as unoccluded ones. 

Cars are more evenly affected by occlusion, and differ from the segmentation model in that unoccluded cars are most consistently detected than partially occluded ones. This is consistent with inspection of the training datasets: KITTI contains many more instances of solitary cars than Cityscapes, which is more often found in clusters.

The low score for nearby, occluded cars (along the bottom of the graph) is largely due to under-representation in Synscapes; there aren't many objects that can occlude a car in the 0-5 meter distance range.

%
%

%
%

\iftrue

\section{Conclusions and future work}

\noindent The goal of this paper is to examine the role of realism in synthetic data. We compared Synscapes to two previous publicly accessible datasets: Synthia, which is purpose-built for the street parsing context, and one generated from Grand Theft Auto (Richter/GTA in the text), which, as a commercial computer game, isn't inherently targeted towards machnine learning, but is entirely street scene-based. The GTA-based dataset is the most varied in terms of number of pedestrian and car models, and contains a large number of different types of buildings and situations. Synscapes has a smaller set of archetypes, but uses much higher fidelity geometry, textures and image synthesis techniques, and also provides entirely unique images.

In order to investigate the role that realism plays, we evaluated
several different uses of synthetic data in the computer vision-based machine learning context. First, we looked at the validation case, which represents the use of virtual simulation to test networks that have been trained on organic, real-world data. Next, we looked at the use of synthetic data as a source for training material and evaluated the resulting networks on organic data. Finally, we studied each synthetic datasets to see how well they can learn to predict on images that are part of their own training domain. 

In each case, and for both semantic segmentation and object detection, Synscapes performs significantly better than the other datasets. As the performance advantages hold consistently across each evaluation method, we see strong evidence for the importance of realism in synthetic data. There are no indications that neural networks are able to naturally abstract away domain shift, and as such it is important for software that aims to achieve accurate sensor simulation attempt to achieve the highest realism possible.

Finally, we also leveraged Synscapes' unique images, annotations and metadata to yield insights into an existing organically-trained neural network. By using the synthetic data as a richly-annotated testing proxy, we were able to yield insights into the Cityscapes and KITTI datasets, and discover correlations and biases in networks trained on them.

Future work could explore in finer detail the difference that individual choices in realism and sensor simulation fidelity makes, from geometric and textural variation and accuracy, to material physicality and illumination complexity. But until such metrics are established, the fundamental indications point to realism playing a large part in making simulation a core of the machine learning and computer vision playbook.

\fi

%
%

{
\small
\bibliographystyle{acm}
\bibliography{egbib}
}

\clearpage

\appendix

\section{Appendix}
\label{sec:appendix}

\subsection{Camera metadata}

\noindent The camera intrinsics and extrinsics are constant throughout the dataset, but for purposes of completeness, they can be found in the \texttt{"camera"} key of the metadata files.

\begin{verbatim}
  "extrinsic": {
    "pitch": 0.038, 
    "roll": -0.0, 
    "x": 1.7, 
    "y": 0.1, 
    "yaw": -0.0195, 
    "z": 1.22
  }, 
  "intrinsic": {
    "fx": 1590.83437, 
    "fy": 1592.79032, 
    "resx": 1440, 
    "resy": 720, 
    "u0": 771.31406, 
    "v0": 360.79945
  }
\end{verbatim}

\subsection{Scenario metadata}

\noindent The variables used to drive the configuration of each image can be found under the \texttt{scene} key. 

\begin{itemize}
\item \texttt{altitude\_variation} specifies the height difference in meter. Note that the extrema may be outside the camera's view.
\item \texttt{curb\_height} in meters.
\item \texttt{ego\_speed} in meters per second. This implicitly indicates the amount of overall motion blur in each image.
\item \texttt{fence\_height} in meters.
\item \texttt{fence\_presence} specifies whether fences are present in the scene. Note that they may be hidden from the camera.
\item \texttt{median\_presence} indicates whether there is a median in the middle of the road.
\item \texttt{num\_*} determine the number of actors of a given class are visible in the image. 
\item \texttt{parking\_angle} defines the angle at which cars are parked; either 0 (parallel), 45 or 90 degrees. Some images do not include a parking lane, as specified by \texttt{parking\_presence}.
\item \texttt{rel\_dist\_to\_isect} contains the distance from the center of the ego vehicle to the center of the next street intersection. 
\item \texttt{sidewalk\_width} in meters.
\item \texttt{sky\_contrast} expressed as the natural logarithm of the ratio of the 99th percentile pixel value and the mean pixel value. Values around 2.0 indicate overcast conditions, with sunny mid-day images around 6.0.
\item \texttt{sun\_height} Normalized angular height of the sun with 0.0 at the horizon and 1.0 at zenith. As the sun's height is simulated for a point away from the Earth's equator, the sun height never reaches full zenith.
\item \texttt{wall\_height} in meters, with \texttt{wall\_presence} indicating whether the scene contains walls.
\end{itemize}

\subsection{Instance metadata}

\noindent Each instance of the non-static classes (person, rider, car, truck, bus, train, motorcycle and bicycle) is annotated with an instance id (see above), and for each instance id the metadata contains information about the following:

\begin{itemize}
\item \texttt{bbox2d} specifies the image bounding box horizontally as $[\texttt{xmin}, \texttt{xmax}]$ and vertically in $[\texttt{ymin}, \texttt{ymax}]$. Both use normalized image coordinates. The depth extents are available in $[\texttt{zmin}, \texttt{zmax}]$, specified in meters. 
\item \texttt{bbox3d} provides an oriented 3D bounding box, defined by the \texttt{origin} (at the rear lower right corner), the \texttt{x} vector facing forward, the \texttt{y} vector to the left, and the \texttt{z} vector facing up. The length of the orientation vectors define the extents in meters. All vectors are defined relative to the ego vehicle reference frame.
\item \texttt{class} indicates the class of the instance for convenience. The same information can be inferred from the \texttt{class} and \texttt{instance} images jointly.
\item \texttt{occluded} specifies the fractional occlusion of each instance, defined as the ratio between actual visible pixels and the number of pixels the instance would occupy if it were completely unoccluded. This is generally accurate to within 1\%.
\item \texttt{truncated} provides the portion of the instance's surface area that lies outside the image view. The accuracy is similarly within 1\%.
\end{itemize}

\end{document}

%% file: table-validation-short.tex
\begin{table*}
\centering
\resizebox{\textwidth}{!}{\begin{tabular}{l|l||l|l|l|l|l|l|l|l|l|l|l|l|l|l|l|l|l|l|l||l}
& Validation 
&  \rotatebox[origin=c]{90}{Road}   
&  \rotatebox[origin=c]{90}{Sidewalk} 
&  \rotatebox[origin=c]{90}{Building} 
&  \rotatebox[origin=c]{90}{Wall}   
&  \rotatebox[origin=c]{90}{Fence}   
&  \rotatebox[origin=c]{90}{Pole}   
&  \rotatebox[origin=c]{90}{Tr.Light} 
&  \rotatebox[origin=c]{90}{Tr.Sign} 
&  \rotatebox[origin=c]{90}{Vegetation} 
&  \rotatebox[origin=c]{90}{Terrain} 
&  \rotatebox[origin=c]{90}{Sky}   
&  \rotatebox[origin=c]{90}{Person} 
&  \rotatebox[origin=c]{90}{Rider}  
&  \rotatebox[origin=c]{90}{Car}    
&  \rotatebox[origin=c]{90}{Truck}  
&  \rotatebox[origin=c]{90}{Bus}    
&  \rotatebox[origin=c]{90}{Train}    
&  \rotatebox[origin=c]{90}{Motorcycle}
&  \rotatebox[origin=c]{90}{Bicycle} 
&  \rotatebox[origin=c]{90}{Mean IoU} \\[20pt]
               \hline 
               
\multirow{4}{*}{FRRN} 

& Cityscapes & 97.11 & 77.78 & 90.2 & 47.22 & 44.54 & 60.67 & 59.72 & 70.44 & 91.08 & 58.06 & 93.02 & 74.85 & 51.01 & 92.23 & 52.51 & 72.32	& 53.49	& 41.61& 	69.26	& 68.27\\

\cline{3-22}

& Synscapes & \textbf{93.69} & \textbf{76.07} & \textbf{84.60} & \textbf{40.98} & \textbf{24.24} & \textbf{47.86} & \textbf{43.54} & \textbf{61.13} & \textbf{88.57} & \textbf{77.62} & \textbf{96.62} & \textbf{65.86} & \textbf{40.60} & \textbf{78.73} & 20.38 & \textbf{12.45} & \textbf{24.22} & \textbf{38.88} & \textbf{46.41} & \textbf{55.92}  \\ 

& Richter  & 82.44 & 31.36 & 71.99 & 28.63 & 12.45 & 0 & 34.71 & 8.2 & 66.12 & 22.91 & 83.58 & 39.01 & 25.7 & 67.97 & \textbf{33.53} & 9.54 & 0.09 & 28.75 & 0 & 34.05 \\

& Synthia & 40.84 & 18.53 & 62.47 & 4.52 & 0.51 & 18.21 & 0.24 & 1.15 & 48.46 & 0 & 80.55 & 21.19 & 4.02 & 27.68 & 0 & 0.99 & 0 & 10.3 & 2.96 & 18.03 \\

\hline

\multirow{4}{*}{DeepLab} 
  

& Cityscapes & 98.15 & 84.85 & 92.71 & 57.34 & 62.14 & 65.20 & 68.62 & 78.87 & 92.68 & 63.46 & 95.33 & 82.26 & 62.84 & 95.37 & 85.30 & 89.10 & 80.92 & 64.55 & 77.34 & 78.79\\

\cline{3-22}


& Synscapes & \textbf{94.83} & \textbf{77.76} & \textbf{88.02} & \textbf{49.66} & \textbf{35.70} & \textbf{59.78} & \textbf{53.90} & \textbf{80.73} & \textbf{89.86} & \textbf{79.73} & \textbf{95.40} & \textbf{78.74} & \textbf{63.19} & 72.68 & 23.64 & 15.76 & \textbf{22.94} & \textbf{62.41} & \textbf{64.49} & \textbf{63.64}\\


& Richter & 86.67 & 43.09 & 82.67 & 39.74 & 24.03 & 0.00 & 46.27 & 30.80 & 67.28 & 23.83 & 90.81 & 71.83 & 54.65 & \textbf{80.34} & \textbf{59.58} & \textbf{16.62} & 0.16 & 52.01 & 0.53 & 45.84 \\


& Synthia & 43.29 & 13.51 & 53.26 & 8.37 & 1.82 & 22.82 & 8.10 & 11.58 & 50.69 & 0.00 & 76.05 & 54.27 & 19.57 & 51.79 & 0.00 & 5.41 & 0.00 & 21.05 & 12.28 & 23.89 \\

\hline

\end{tabular}}
\caption{Validation on synthetic data for reference versions of FRRN and DeepLab v3+ architectures. 
}
\label{table:validation_results}
\end{table*}

%% file: table-det-validation1.tex
\begin{table}
\centering
\resizebox{\linewidth}{!}{
\begin{tabular}{l|l|c|c|c}
Training & Validation & mAP & mAP@0.50 &  mAP@0.75\\
\hline
KITTI & KITTI & 0.456 & 0.716 & 0.484  \\
KITTI & GTA & 0.061& 0.115& 0.059 \\
KITTI & Synscapes & 0.206 & 0.400 & 0.187 \\
KITTI + Synscapes  & Synscapes & 0.570 & 0.813 & 0.634 \\
\hline
\end{tabular}}
\caption{COCO metric results for object detection using Faster-RCNN with Resnet101 from Google's Tensorflow Object Detection API. The mean AP (mAP) is computed for IoU [0.5 : 0.95], and mAP@0.50 and mAP@0.75 are computed for IoU 0.50 and 0.75 respectively.}
\label{table:validation_det_1}
\end{table}

%% file: table-det-validation2.tex
\begin{table}
{\footnotesize
\centering
\resizebox{\linewidth}{!}{
\begin{tabular}{l|l|c|c}

Training & Validation & Car & Pedestrian\\
\hline
SynScapes & SynScapes & 0.752 & 0.802 \\
KITTI & KITTI & 0.534 & 0.847 \\
KITTI & GTA & 0.021 & 0.096 \\
KITTI  & Synscapes & 0.355 & 0.438\\
KITTI + Synscapes & Synscapes & 0.843 & 0.782 \\  
\hline
\end{tabular}}
}
\caption{Individual scores for classes Car and Pedestrian from Faster-RCNN. Average precision (AP) is computed at IoU = 0.50 using the Pascal VOC metric.}
\label{table:validation_det2}

\end{table}

%% file: table-training.tex
\begin{table*}
\centering
\resizebox{\textwidth}{!}{\begin{tabular}{l|l||l|l|l|l|l|l|l|l|l|l|l|l|l|l|l|l|l|l|l||l}
& Training 
&  \rotatebox[origin=c]{90}{Road}   
&  \rotatebox[origin=c]{90}{Sidewalk} 
&  \rotatebox[origin=c]{90}{Building} 
&  \rotatebox[origin=c]{90}{Wall}   
&  \rotatebox[origin=c]{90}{Fence}   
&  \rotatebox[origin=c]{90}{Pole}   
&  \rotatebox[origin=c]{90}{Tr.Light} 
&  \rotatebox[origin=c]{90}{Tr.Sign} 
&  \rotatebox[origin=c]{90}{Vegetation} 
&  \rotatebox[origin=c]{90}{Terrain} 
&  \rotatebox[origin=c]{90}{Sky}   
&  \rotatebox[origin=c]{90}{Person} 
&  \rotatebox[origin=c]{90}{Rider}  
&  \rotatebox[origin=c]{90}{Car}    
&  \rotatebox[origin=c]{90}{Truck}  
&  \rotatebox[origin=c]{90}{Bus}    
&  \rotatebox[origin=c]{90}{Train}    
&  \rotatebox[origin=c]{90}{Motorcycle}
&  \rotatebox[origin=c]{90}{Bicycle} 
&  \rotatebox[origin=c]{90}{Mean IoU} \\[20pt]
               \hline 
               
\multirow{7}{*}{FRRN} 

& Cityscapes 
& 97.11 & 77.78 & 90.2 & 47.22 & 44.54 & 60.67 & 59.72 & 70.44 & 91.08 & 58.06 & 93.02 & 74.85 & 51.01 & 92.23 & 52.51 & 72.32	& 53.49	& 41.61& 	69.26	& 68.27 \\ 

\cline{2-22}

& SynScapes 
& \textbf{90.71} & \textbf{47.54} & \textbf{76.46} & \textbf{23.16} & \textbf{25.12} & \textbf{37.19} & \textbf{35.4} & \textbf{38.95} & \textbf{79.11} & \textbf{20.87} & \textbf{86.4} & \textbf{58.56} & \textbf{39.72} & \textbf{82.09} & 14.48 & \textbf{27.84} & \textbf{14.7} & \textbf{12.91} & \textbf{47.53} & \textbf{45.20} \\ 

& Richter 
& 40.27 & 21.15 & 62.52 & 7.16 & 6.85 & 8.41 & 11.03 & 1.52 & 75.42 & 12.55 & 60.09 & 31.71 & 0 & 27.43 & \textbf{14.91} & 7.47 & 7.98 & 0.23 & 0.02 & 20.88\\

& Synthia 
& 60.8 & 28.05 & 59.86 & 0.07 & 0.07 & 25.52 & 2.34 & 2.69 & 74.62 & 0 & 75.04 & 38.33 & 3.84 & 35.81 & 0 & 2.09 & 0 & 1.92 & 2.74 & 21.78 \\

\cline{2-22}

& SynScapes + CS 
& \textbf{97.70} & \textbf{81.64} & \textbf{91.27} & \textbf{51.34} & 49.37 & \textbf{65.35} & \textbf{66.87} & \textbf{75.59} & \textbf{91.78} & \textbf{61.09} & \textbf{94.15} & \textbf{78.65} & \textbf{58.22} & \textbf{93.94} & \textbf{70.51} & \textbf{82.40} & \textbf{79.09} & \textbf{54.30} & \textbf{72.63} & \textbf{74.52} \\ 

& Richter + CS 
& 96.90 & 77.17 & 90.71 & 49.20 & 48.62 & 62.42 & 61.58 & 72.34 & 91.25 & 60.93 & 93.84 & 75.53 & 53.77 & 93.64 & 64.19 & 73.13 & 61.44 & 46.80 & 70.96 & 70.76 \\

& Synthia + CS 
& 97.58 & 81.04 & 90.81 & 47.58 & \textbf{50.49} & 62.48 & 63.05 & 73.45 & 91.47 & 60.39 & 93.80 & 77.11 & 53.05 & 93.19 & 57.04 & 73.21 & 52.64 & 38.07 & 71.51 & 69.89 \\

\hline

\multirow{7}{*}{DeepLab} 

& Cityscapes 
& 97.98 & 83.88 & 92.18 & 59.36 & 59.14 & 61.69 & 65.60 & 75.70 & 92.11 & 60.74 & 94.44 & 80.53 & 58.41 & 94.57 & 81.31 & 85.87 & 78.42 & 58.88 & 73.87 & 76.56 \\ 

\cline{2-22}

& Synscapes 
& \textbf{85.38} & \textbf{47.75} & \textbf{73.87} & \textbf{27.75} & \textbf{31.83} & \textbf{46.89} & \textbf{50.14} & \textbf{58.65} & \textbf{85.92} & \textbf{41.12} & \textbf{83.87} & \textbf{66.38} & \textbf{26.98} & \textbf{84.01} & \textbf{25.95} & 18.99 & \textbf{5.17} & \textbf{35.04} & \textbf{61.00} & \textbf{50.35} \\ 

& Richter 
& 54.87 & 24.74 & 50.15 & 15.64 & 9.24 & 39.21 & 35.55 & 13.54 & 81.12 & 27.75 & 38.07 & 58.48 & 17.34 & 78.63 & 23.51 & \textbf{31.07} & 0.28 & 12.69 & 0.00 & 32.20 \\ 

& Synthia 
& 71.04 & 29.91 & 69.55 & 3.24 & 0.15 & 33.09 & 28.89 & 12.46 & 76.22 & 0.00 & 71.95 & 65.98 & 23.21 & 75.92 & 0.00 & 23.82 & 0.00 & 13.13 & 23.25 & 32.73\\ 

\cline{2-22}

& Synscapes + CS 
& \textbf{98.14} & \textbf{84.82} & \textbf{92.95} & 57.66 & 62.52 & \textbf{66.23} & \textbf{70.06} & \textbf{78.76} & \textbf{92.72} & 61.32 & \textbf{95.05} & \textbf{82.86} & \textbf{62.92} & \textbf{95.46} & \textbf{84.59} & \textbf{88.98} & \textbf{79.21} & \textbf{65.95} & \textbf{77.94} & \textbf{78.85} \\ 

& Richter + CS 
&  97.45 & 81.46 & 92.66 & \textbf{58.45} & \textbf{63.31} & 65.23 & 69.37 & 78.69 & 92.49 & \textbf{62.60} & 94.93 & 81.79 & 60.96 & 95.20 & 80.99 & 85.54 & 72.18 & 63.71 & 76.80 & 77.57 \\ 

& Synthia + CS 
& 97.98 & 84.12 & 92.52 & 54.33 & 56.56 & 64.80 & 68.84 & 78.01 & 92.51 & 61.41 & 95.00 & 82.05 & 62.60 & 95.14 & 81.58 & 86.28 & 77.32 & 63.42 & 76.99 & 77.45 \\ 

\hline

\hline

\end{tabular}}
\caption{Base training and fine tuning results for FRRN and Deeplab v3+ architectures.}
\label{table:finetuning_results}
\end{table*}

%% file: table-validation.tex
\begin{table*}
\centering
\resizebox{\textwidth}{!}{\begin{tabular}{l|l|l||l|l|l|l|l|l|l|l|l|l|l|l|l|l|l|l|l|l|l||l}
& Training & Validation 
&  \rotatebox[origin=c]{90}{Road}   
&  \rotatebox[origin=c]{90}{Sidewalk} 
&  \rotatebox[origin=c]{90}{Building} 
&  \rotatebox[origin=c]{90}{Wall}   
&  \rotatebox[origin=c]{90}{Fence}   
&  \rotatebox[origin=c]{90}{Pole}   
&  \rotatebox[origin=c]{90}{Tr.Light} 
&  \rotatebox[origin=c]{90}{Tr.Sign} 
&  \rotatebox[origin=c]{90}{Vegetation} 
&  \rotatebox[origin=c]{90}{Terrain} 
&  \rotatebox[origin=c]{90}{Sky}   
&  \rotatebox[origin=c]{90}{Person} 
&  \rotatebox[origin=c]{90}{Rider}  
&  \rotatebox[origin=c]{90}{Car}    
&  \rotatebox[origin=c]{90}{Truck}  
&  \rotatebox[origin=c]{90}{Bus}    
&  \rotatebox[origin=c]{90}{Train}    
&  \rotatebox[origin=c]{90}{Motorcycle}
&  \rotatebox[origin=c]{90}{Bicycle} 
&  \rotatebox[origin=c]{90}{Mean IoU} \\[20pt]
               \hline

\multirow{4}{*}{DeepLab v3+} 
  

& Cityscapes & Cityscapes & 98.15 & 84.85 & 92.71 & 57.34 & 62.14 & 65.20 & 68.62 & 78.87 & 92.68 & 63.46 & 95.33 & 82.26 & 62.84 & \textbf{95.37} & \textbf{85.30} & 89.10 & 80.92 & 64.55 & \textbf{77.34} & 78.79\\


& Synscapes & SynScapes & \textbf{98.58} & \textbf{93.80} & \textbf{95.20} & \textbf{92.75} & \textbf{87.48} & \textbf{72.50} & \textbf{74.28} & \textbf{86.78} & \textbf{93.20} & \textbf{90.38} & \textbf{97.24} & \textbf{84.52} & \textbf{77.87} & 93.99 & 85.10 & 88.48 & \textbf{89.87} & 76.46 & 74.57 & \textbf{87.00} \\

& Richter & Richter & 97.14 & 85.71 & 90.42 & 63.70 & 45.01 & 0.00 & 65.14 & 61.56 & 86.17 & 74.79 & 96.68 & 76.36 & 39.97 & 90.05 & 84.15 & 82.35 & 0.02 & 58.75 & 0.00 & 63.05 \\


& Synthia & Synthia & 93.14 & 91.27 & 92.02 & 69.33 & 57.60 & 54.43 & 17.90 & 31.78 & 79.51 & 0.00 & 93.96 & 74.64 & 53.05 & 89.28 & 0.00 & \textbf{89.85} & 0.00 & 69.77 & 29.74 & 57.22 \\

\hline

\end{tabular}}
\caption{Self-validation results for Cityscapes (reference model), Synscapes, Richter and Synthia using DeepLab v3+. 
}
\label{table:cross_validation_results}
\end{table*}

%% file: table-det-training1.tex
\begin{table}
\centering
\resizebox{\linewidth}{!}{
\begin{tabular}{l|l|c|c|c}
Training & Validation & mAP & mAP@0.50 &  mAP@0.75\\
\hline
KITTI & KITTI & 0.456 & 0.716 & 0.484 \\
Synscapes & KITTI & 0.092 &  0.237  & 0.069  \\
Synscapes + KITTI & KITTI & 0.519 &  0.791 & 0.589 \\
Synscapes & Synscapes & 0.488 & 0.777 & 0.518 \\
\hline
\end{tabular}}
\caption{COCO metric results for object detection using Faster-RCNN with Resnet101 from Google's Tensorflow Object Detection API. The mean AP (mAP) is computed for IoU [0.5 : 0.95], and mAP@0.50 and mAP@0.75 are computed for IoU 0.50 and 0.75 respectively.}
\label{table:training_det1}
\end{table}

%% file: table-det-training2.tex
\begin{table}
\centering
\begin{tabular}
{l|l|c|c}
Training & Validation & Car & Pedestrian\\
\hline
KITTI & KITTI & 0.534 & 0.847 \\
SynScapes & KITTI & 0.344 & 0.131 \\
SynScapes + KITTI & KITTI  & 0.902 & 0.679\\
\hline
\end{tabular}
\caption{Individual results for classes 'car' and 'pedestrian' for average precision (AP) computed at IoU = 0.50 using the Pascal VOC metric. MW: KittiBox or RCNN?}
\label{table:training_det2}
\end{table}

%% file: table-det-training3.tex
\begin{table}
\centering
\begin{tabular}
{l|c|c|c}
Training & Easy & Medium & Hard \\
\hline
KITTI &  97.4\% & 89.0\% & 75.1 \%\\
SynScapes & 68.6\% & 53.4\% & 44.5 \%\\
SynScapes + KITTI & 98.7\% & 90.0\% & 77.0\%\\
\hline
\end{tabular}
\caption{Validation results from FastBox~\cite{DBLP:journals/corr/TeichmannWZCU16}, using the KittiBox implementation. The training was performed from scratch and evaluated on 500 test images extracted from the KITTI dataset. The result scores are given as the average precision (AP) for the 'car' and 'pedestrian' classes. }
\label{table:training-det3}
\end{table}